\documentclass{article} 
\PassOptionsToPackage{table}{xcolor}
\usepackage{iclr2027_conference,times}

\usepackage{amsmath,amsfonts,bm}

\def\eqref#1{equation~\ref{#1}}

\def\1{\bm{1}}

\DeclareMathAlphabet{\mathsfit}{\encodingdefault}{\sfdefault}{m}{sl}
\SetMathAlphabet{\mathsfit}{bold}{\encodingdefault}{\sfdefault}{bx}{n}

\usepackage[table]{xcolor}
\usepackage{booktabs}
\usepackage{graphicx}
\usepackage{multirow}
\usepackage{makecell}
\usepackage{wrapfig}
\usepackage{amssymb}      
\usepackage{enumitem}
\usepackage[T1]{fontenc}
\usepackage[type1]{josefin}
\usepackage{pifont}
\usepackage{algorithm}
\usepackage{algpseudocode}

\usepackage{hyperref}
\usepackage{url}
\usepackage{caption} 

\definecolor{GoodGreen}{HTML}{2E7D32} 
\definecolor{BadRed}{HTML}{C62828}    
\newcommand{\dui}{\textcolor{GoodGreen}{\ding{51}}}
\newcommand{\cuo}{\textcolor{BadRed}{\ding{55}}}
\definecolor{GroupGray}{RGB}{235,235,235}
\definecolor{OursTint}{RGB}{230,243,255}

\title{\textbf{Rolling Sink}: Bridging Limited-Horizon Training and Open-Ended Testing in Autoregressive Video Diffusion}

\author{
Haodong Li$^1$\ \ Shaoteng Liu$^2$\ \ Zhe Lin$^2$\ \ Manmohan Chandraker$^1$
\\
$^1$UCSD\ \ $^2$Adobe
\\
\texttt{\small\{hal211,mkchandraker\}@ucsd.edu; \{shaotengl,zlin\}@adobe.com}
}

\iclrfinalcopy 
\begin{document}

\maketitle

\begin{center}
\vspace{-.5cm}
    \captionsetup{type=figure}
    \includegraphics[width=\linewidth]{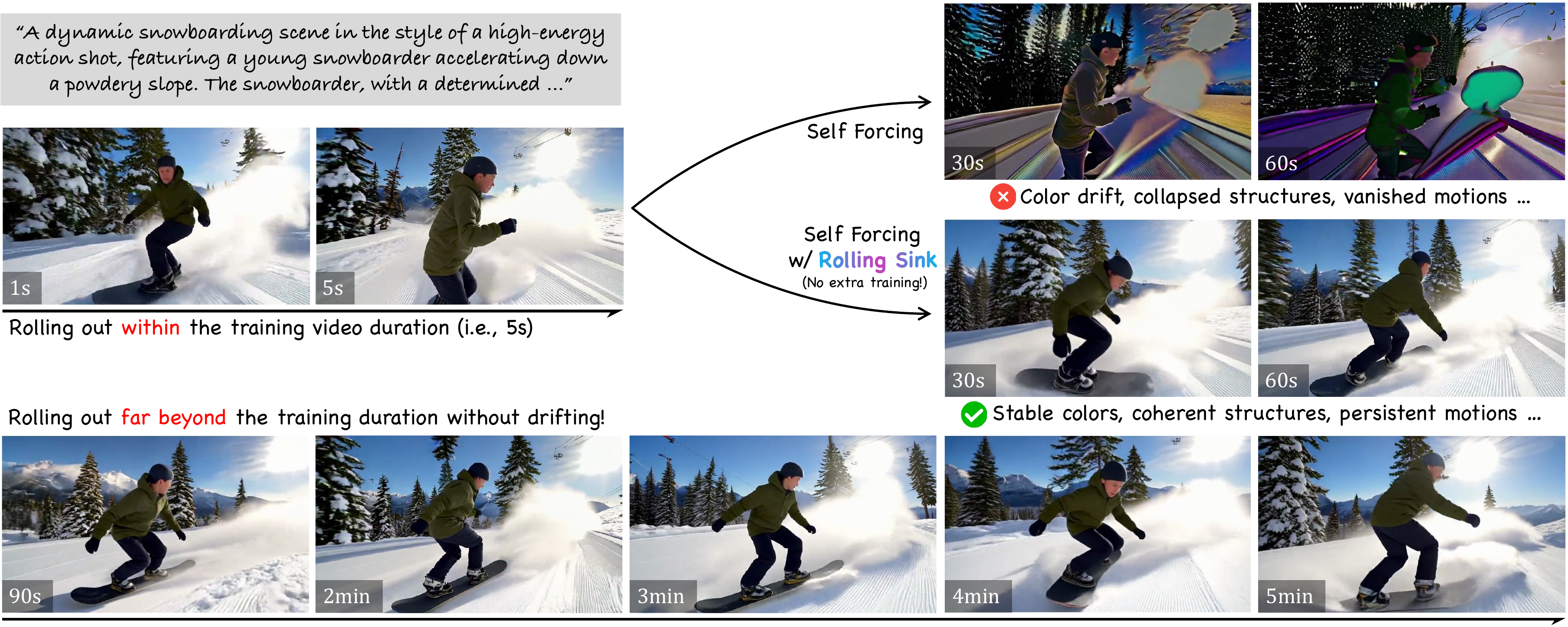}
    \captionof{figure}{
    \textbf{Rolling Sink extends a short-horizon-trained AR video diffusion model to much longer test-time rollouts.}
    Without extra training, it keeps a 5s-trained model rolling far beyond its training horizon under a bounded cache: a 5-minute rollout here, and a 30-minute qualitative example in Fig.~\ref{supp:fig:30min} (App.~\ref{supp:sec:30min}).
    }
    \label{fig:teaser}
\end{center}

\begin{abstract}
Autoregressive (AR) video diffusion models in the Self Forcing family generate a video chunk by chunk, each conditioned on the ones before it, so they can keep rolling out with no preset length.
In practice they are trained on short, fixed clips (\emph{e.g.}, 5s).
Once a rollout passes this training horizon, the cache fills with self-generated chunks the model never saw in training, and quality degrades: colors oversaturate, identities drift, structures collapse, and motion fades.
We frame this failure as a \emph{cache-behavior mismatch}, a domain shift between the within-horizon caches seen in training and those produced by open-ended testing, and address it with \textbf{Rolling Sink}, a training-free\footnote{Training-free means no additional training on top of the base AR model, which is itself trained. Because open-ended testing can exceed any finite training duration, this gap cannot be fully closed by training on longer clips alone, which motivates an inference-time solution.\label{fn:trainingfree}} inference-time cache policy.
Within the base model's fixed cache budget, Rolling Sink holds the cache close to its stable within-horizon behavior: it restores low-drift content, sliding temporal indices, and evolving semantics, and keeps a small recent context.
Attention sink and temporal re-indexing are existing mechanisms. Our contribution is the mismatch formulation, the joint bounded-cache policy, and the Rolling Semantics schedule.
On two 5s-trained bases, Self Forcing and Causal Forcing, Rolling Sink attains the best averaged rank on \texttt{VBench-Long} at both 1- and 5-minute rollouts under the same inference setting, at about 0.56\% throughput change, and holds GPU memory constant.
It also sustains a 30-minute qualitative rollout, $360\times$ its training horizon.
We validate at 1- and 5-minute rollouts and do not claim guaranteed quality at arbitrary duration.
Project page: \url{https://rolling-sink.github.io/}.
\end{abstract}

\section{Introduction}
\label{sec:intro}

Shot length in film is open-ended: a take may last seconds or run for minutes, like the 16.5-minute opening dialogue of Steve McQueen's \emph{Hunger}~\citep{mcqueen2008hunger} or the minute-long tricycle shot in Kubrick's \emph{The Shining}~\citep{kubrick1980shining}.
This motivates open-ended generation, where the length is not fixed and the model must keep producing coherent frames for as long as asked.

Large video diffusion models~\citep{sora,wan2025,kling2.6,veo3,gen4,seedance2026seedance} denoise all frames at once, which ties them to short, fixed clips and rules out this setting.
AR video diffusion models in the Self Forcing family~\citep{huang2025self,zhu2026causal,yang2025longlive,liu2025rolling} instead predict each chunk from previously generated ones, which lets them keep rolling out with no preset length.
These models are trained on short clips (\emph{e.g.}, 5s).
Within the training horizon they are stable, but once a rollout passes it they degrade fast: oversaturated colors, collapsed structures, inconsistent identities, and vanishing motion (Fig.~\ref{fig:teaser}).

We read this degradation as a domain shift on the AR cache.
The prompt embedding is fixed, and each chunk starts from the same noise distribution. The only input that changes across the rollout is thus the cache, \emph{i.e.}, past self-generated chunks.
Within the horizon, the cache stays in the training distribution, so the model behaves as trained.
Beyond the horizon, the cache falls outside it. Each prediction is then biased, and the errors compound over the rollout, an instance of exposure bias~\citep{bengio2015scheduled,lamb2016professor}.
Limited-horizon training cannot cover open-ended lengths, leaving the cache as the only test-time lever.
We call this gap a \emph{cache-behavior mismatch} and reduce it by making the test-time cache resemble the within-horizon cache seen in training, along the three factors the model conditions on: low-drift content, sliding temporal indices, and evolving semantics.

Building on this view, we propose \textbf{Rolling Sink}, a training-free\textsuperscript{\ref{fn:trainingfree}} inference-time cache policy that runs under the base model's fixed budget and few-step denoising, keeping the base's streaming efficiency (App.~\ref{supp:sec:fps}).
To restore content, it pins a few \emph{sink} chunks generated within the training horizon, following the attention-sink idea from streaming language models~\citep{xiao2023efficient}, and re-indexes them to the sliding window before the current chunk, as in prior horizon-extension work~\citep{yi2025deep,yesiltepe2025infinity}.
To restore evolving semantics, \emph{Rolling Semantics} traverses the finite bank of low-drift chunks in a forward-reverse schedule and reads a sliding window from it, while one retained recent chunk anchors local motion.

Attention sinks and temporal re-indexing are existing mechanisms, so we do not claim them as contributions.
Rolling Sink contributes the framing and the policy that combines them: (i) casting the horizon gap as a cache-behavior mismatch, (ii) a bounded-cache policy that jointly maintains low-drift content, sliding temporal indices, and evolving semantics, and (iii) the Rolling Semantics schedule over a finite low-drift bank.
This sets Rolling Sink apart from prior methods that mainly target a single axis: positional extrapolation~\citep{yesiltepe2025infinity,li2026train} or context retention~\citep{yi2025deep,kim2026memrope}.
Tab.~\ref{tab:method} gives the per-method comparison.

We evaluate on \texttt{VBench-Long}~\citep{huang2023vbench,huang2025vbench++,zheng2025vbench2} over 1- and 5-minute rollouts, with qualitative rollouts in Fig.~\ref{fig:qc} and a 30-minute example in App.~\ref{supp:sec:30min}.
On two 5s-trained bases, Self Forcing~\citep{huang2025self} and Causal Forcing~\citep{zhu2026causal}, Rolling Sink attains the best averaged rank at both lengths under the same inference setting, and its Self Forcing variant ranks first even against training-based baselines, including LongLive~\citep{yang2025longlive} trained on 1-minute clips.
Averaged rank aggregates over all \texttt{VBench-Long} dimensions, so Rolling Sink does not lead on every one. Sec.~\ref{sec:exp} reports the per-dimension trade-offs.
Rolling Sink keeps generating at constant GPU memory.
The 1- and 5-minute numbers are quantitative, the 30-minute result is qualitative, and we do not claim guaranteed quality at arbitrary duration.

In summary, our contributions are:
\ding{172} \textbf{Formulation.} We frame long-horizon AR drift as a cache-behavior mismatch between limited-horizon training and open-ended testing, and identify bounded-cache maintenance as the lever for narrowing it.
\ding{173} \textbf{Method.} We introduce Rolling Sink, a training-free bounded-cache policy that jointly maintains low-drift content, sliding temporal indices, and evolving semantics, with Rolling Semantics as its one new component.
\ding{174} \textbf{Evidence.} Rolling Sink attains the best averaged rank on \texttt{VBench-Long} at 1- and 5-minute rollouts on two 5s-trained bases, at about 0.56\% throughput change, and sustains a 30-minute qualitative rollout.

\section{Related Work}
\label{sec:rw}

\subsection{Autoregressive Video Diffusion}
Large video diffusion models~\citep{blattmann2023stable,yang2024cogvideox,kong2024hunyuanvideo,sora,wan2025,veo3,gen4,seedance2026seedance,zhang2026dvd} build on image diffusion~\citep{ho2020denoising,rombach2022high,song2020denoising,he2024disenvisioner,he2024lotus,he2025lotus} and denoise all frames jointly with bidirectional attention~\citep{peebles2023scalable}, which ties them to short, fixed clips.
AR video diffusion instead factorizes a video into chunk-wise conditionals and predicts each chunk from prior ones~\citep{chen2024diffusion,deng2024autoregressive,yin2024one,yin2025slow,teng2025magi,chen2025skyreels,guo2025long,zhang2025frame,ren2025next,zhou2026videomemory}, in principle enabling unbounded rollout.
Self Forcing~\citep{huang2025self} aligns the within-horizon cache by training on the model's own generations, and Causal Forcing~\citep{zhu2026causal}, LongLive~\citep{yang2025longlive}, and Rolling Forcing~\citep{liu2025rolling} extend the recipe with better distillation or longer rollouts.
All of these modify training yet still train on short clips (5--6.5s, up to 1 minute for LongLive), so they drift once the rollout exceeds the trained length (Fig.~\ref{fig:qc}).
Rolling Sink leaves training untouched and controls the cache at inference.

\subsection{Training-Free Test-Time Horizon Extension}
\label{sec:rw:tf}
A second line keeps the base model frozen and edits the cache at test time.
One group corrects generation dynamics: $\infty$-RoPE~\citep{yesiltepe2025infinity} and FLEX~\citep{li2026train} extend temporal positional indices past the trained range, and Pathwise TTC~\citep{xiang2026pathwise} corrects sampling trajectories against a stable reference.
These target positional extrapolation and do not restore low-drift content once recent chunks drift.
Another group manages historical context under bounded memory.
Deep Forcing~\citep{yi2025deep} keeps persistent deep-sink chunks and realigns their RoPE phases to the current timeline, MemRoPE~\citep{kim2026memrope} pairs memory tokens with online RoPE indexing, and Head Forcing~\citep{tian2026head}, Echo Forcing~\citep{wu2026echo}, and Pack Forcing~\citep{mao2026packforcing} adopt head-wise, scene-level, or hierarchical cache policies.
These retain a low-drift or memory prefix but pin its semantics, so the retained context stops evolving with the rollout.
The attention-sink and RoPE re-indexing mechanisms that Rolling Sink reuses originate here and in streaming language models~\citep{xiao2023efficient}, which keep initial sink tokens beside a recent window under relative positions.
Rolling Sink instead matches the cache to its within-horizon behavior along three factors at once: low-drift content, sliding temporal indices, and evolving semantics.
Among the compared methods, it is the only training-free policy to maintain all three under the base model's bounded budget.
Its new component, Rolling Semantics, keeps the retained low-drift context evolving in semantics, the axis the retention methods above leave fixed.
Tab.~\ref{tab:method} summarizes where each method sits.

\section{Methodology}

\begin{figure*}[!t]
    \centering
    \includegraphics[width=\linewidth]{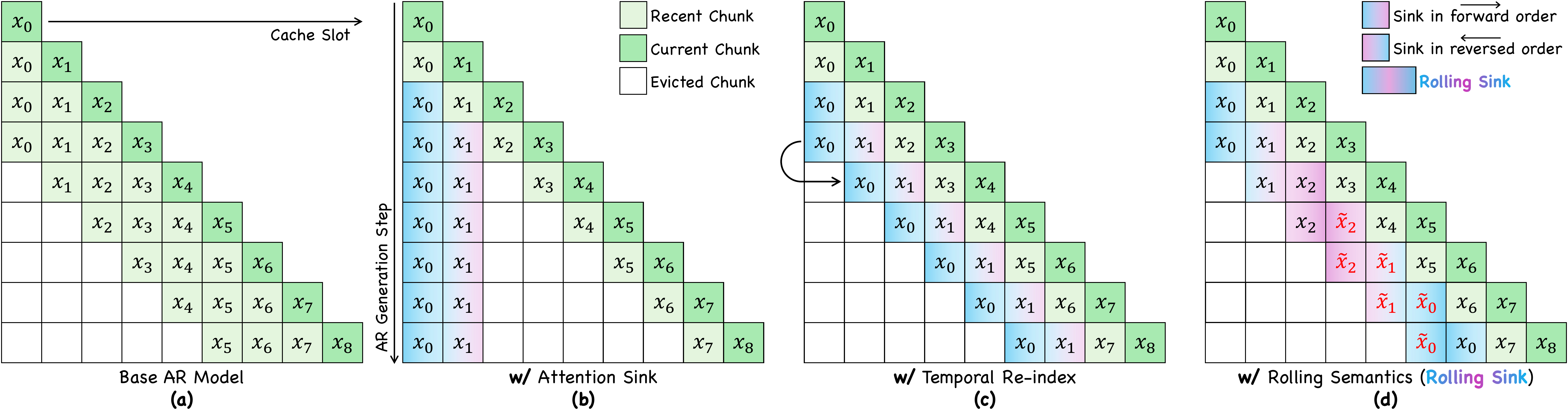}
\vspace{-.5cm}
\caption{
\textbf{Cache analysis and derivation of Rolling Sink.}
\textbf{(a)} The base AR model uses a bounded recent cache.
Tested beyond the training horizon, its temporal indices still slide and its semantics still evolve with the rollout, but the cached content gradually drifts.
\textbf{(b)} Attention Sink reuses low-drift chunks generated within the training horizon, restoring low-drift content but leaving the sink chunks fixed in temporal indices and semantics.
\textbf{(c)} Temporal Re-index assigns these sink chunks to the recent window just before the current chunk, restoring sliding temporal indices.
\textbf{(d)} Rolling Semantics traverses the low-drift chunks in a forward-reverse schedule, then reads a sliding semantic window from this traversal to restore evolving semantics.
One recent chunk is retained to keep local motion smooth.
Under a bounded budget, Rolling Sink keeps the within-horizon cache behavior during long rollouts, narrowing the gap between limited-horizon training and open-ended testing.
}
    \label{fig:method}
\vspace{-.3cm}
\end{figure*}

\subsection{Preliminaries}
\label{sec:pre}

An AR video diffusion model generates a video as a sequence of $N$ chunks $\boldsymbol{x}_{[0,N)}=(\boldsymbol{x}_0,\dots,\boldsymbol{x}_{N-1})$, denoising each from the prior ones:
\begin{equation}
\small
p_\theta\left(\boldsymbol{x}_{[0,N)}\right)
= \prod_{i=0}^{N-1} p_\theta(\boldsymbol{x}_i \mid \boldsymbol{x}_{[0,i)}).
\end{equation}

The cache holds the conditioning context $\boldsymbol{x}_{[0,i)}$, and how it is built during training defines the recipe.
Teacher Forcing~\citep{gao2024ca2,hu2024acdit,jin2024pyramidal,zhang2025test} and Diffusion Forcing~\citep{chen2024diffusion,yin2025slow,teng2025magi,song2025history} condition on clean or noised ground-truth (GT) frames. At test time no GT is available, so the cache instead holds the model's own generations, opening a train-test gap.
Self Forcing~\citep{huang2025self,cui2025self,yin2024improved} and its successors~\citep{zhu2026causal,yang2025longlive,liu2025rolling} train on the model's own generations and close this gap, but only within a finite training horizon.

\subsection{Key Issue: Keeping the Test Cache In-Distribution}
\label{sec:key}

The AR conditional $p_\theta(\boldsymbol{x}_i\mid\mathcal{C})$ is trained only on caches $\mathcal{C}$ from within-horizon rollouts.
Beyond the horizon the cache is out-of-distribution, so the degradation is the ordinary cost of running a model off its training distribution, with exposure bias~\citep{bengio2015scheduled,lamb2016professor} as one instance.
We cannot train on unbounded lengths, so the only test-time lever is to shrink this gap: make the test cache resemble the caches the model saw in training.

The cache enters the model through attention over its chunks. Each chunk thus contributes three things the model reads: content (the attended latent values), temporal position (the RoPE index), and, across chunks, a semantic trajectory.
Matching the training distribution therefore reduces to matching these three factors: \ding{172}~\textbf{low-drift content}, chunks that look like valid within-horizon frames; \ding{173}~\textbf{sliding temporal indices}, the same recent window of positions as in training; and \ding{174}~\textbf{evolving semantics}, a trajectory that moves forward as in a within-horizon rollout.
These are the factors the conditional depends on, not hand-picked properties.
Beyond its horizon a frozen base keeps the last two but drifts on the first, which pushes the whole cache off-distribution.

We restore the three factors in order under the bounded budget of $K$ chunks, content first: matching positions or trajectory over corrupted chunks does not help (Sec.~\ref{sec:method_rs}).
Fig.~\ref{fig:method} shows the four stages, and Tab.~\ref{tab:method} places them against prior methods.

\setlength{\cmidrulewidth}{\lightrulewidth}

\begin{table*}[!t]
\centering
\scriptsize
\setlength{\tabcolsep}{5.3pt}
\caption{
\textbf{Cache mechanism comparison of long-horizon AR methods.}
We view the horizon gap as a domain shift on the AR cache.
The base is trained only on within-horizon caches, so keeping the test cache in-distribution means matching the three factors the model conditions on: content, temporal position, and semantic trajectory.
The first row is the frozen base with no test-time policy.
The other rows are training-free policies applied to it under the same bounded cache.
Cells restate each method's reported mechanism (\emph{drifted}: off-distribution content, \emph{pinned}: static sink, \emph{rolling}: evolving sink).
Attention sink (low-drift content) and RoPE re-indexing (temporal position) come from streaming language models~\citep{xiao2023efficient} and prior training-free horizon-extension work~\citep{yi2025deep,yesiltepe2025infinity}.
But only Rolling Sink matches all three factors, and Rolling Semantics (\emph{rolling}) is its one new component.
}
\vspace{-.1cm}
\label{tab:method}
\begin{tabular}{lcccc}
\toprule
\multirow{2}{*}{Method}
& \multirow{2}{*}{\makecell{Training-\\free}}
& \multicolumn{3}{c}{Cache factors beyond the training horizon} \\
\cmidrule(lr){3-5}
& & Content & Temporal position & Semantic trajectory \\
\midrule
Self / Causal Forcing~\citep{huang2025self,zhu2026causal}
& \cuo & drifted & sliding & drifted \\
$\infty$-RoPE / FLEX~\citep{yesiltepe2025infinity,li2026train}
& \dui & drifted & extrapolated & drifted \\
Deep Forcing~\citep{yi2025deep}
& \dui & low-drift & re-indexed & pinned \\
MemRoPE~\citep{kim2026memrope}
& \dui & low-drift & re-indexed & pinned \\
\rowcolor{OursTint}
\textbf{Rolling Sink} (ours)
& \dui & low-drift & re-indexed & rolling \\
\bottomrule
\end{tabular}
\vspace{-.2cm}
\end{table*}

\subsection{Rolling Sink}
\label{sec:method_rs}

We recover the three factors of Sec.~\ref{sec:key} in order under the bounded budget.
Attention Sink (Sec.~\ref{sec:attnsink}) reuses the attention-sink mechanism~\citep{xiao2023efficient} to restore low-drift content but leaves the sink static. Temporal Re-index (Sec.~\ref{sec:re_idx}) applies the RoPE realignment of prior work~\citep{yi2025deep,yesiltepe2025infinity} to restore sliding temporal indices. Rolling Semantics (Sec.~\ref{sec:roll_sem}), new to this paper, restores evolving semantics.
The first two adapt existing mechanisms. The contribution is combining all three into one bounded-cache policy plus the Rolling Semantics schedule.

\subsubsection{Attention Sink: Low-Drift Content}
\label{sec:attnsink}

We start from a base AR model, where each AR step generates a chunk $\boldsymbol{x}_i$ of $n$ latent frames, $\boldsymbol{x}_i[k]=\boldsymbol{z}_{in+k}$ for $k\in[0,n)$\footnote{Throughout this paper, all interval values are integers unless otherwise specified, \emph{e.g.}, $[0,n)$ indicates $\{0,1,\cdots,n-1\}$.}, where $\boldsymbol{z}$ denotes a latent frame\footnote{Each latent frame corresponds to multiple RGB frames~\citep{wan2025}.}.
Within the training horizon the context is the full prefix $\boldsymbol{x}_{[0,i)}$.
Beyond it, the base keeps only a bounded recent cache of size $K$:
\begin{equation}
\small
p_\theta\left(\boldsymbol{x}_i\mid\boldsymbol{x}_{\left[i-K,i\right)}\right),
\ i\in[K,\infty).
\end{equation}
Its temporal indices slide and its semantics evolve with the rollout, but its chunks gradually drift, degrading quality once out-of-distribution (i.e., beyond the training horizon).

Chunks made within the training horizon are usually low-drift, since the model was optimized in that regime.
To restore low-drift cached content, we reuse the attention-sink mechanism~\citep{xiao2023efficient} and reserve the first $S$ cache slots for a static set of early within-horizon chunks, which we call \emph{sink chunks}:
\begin{equation}
\label{eq:as}
\small
p_\theta\left(
\boldsymbol{x}_i
\mid
\texttt{Cat}\left(
\boldsymbol{x}_{[0,S)},
\boldsymbol{x}_{[i-(K-S),i)}
\right)
\right),
\ i\in[K,\infty),
\end{equation}
where $S\in[0,K]$ is the sink size and $\texttt{Cat}(\cdot)$ concatenates the sink and recent chunks.
This restores low-drift content, but leaves the sink chunks static in both temporal indices and semantics.

\subsubsection{Temporal Re-index: Sliding Temporal Indices}
\label{sec:re_idx}

The sink chunks keep their original temporal indices throughout the rollout, unlike the within-horizon cache, where chunks occupy a sliding window just before the current chunk.
We re-index the sink chunks to this sliding window while keeping their low-drift content unchanged.

Specifically, we use $\boldsymbol{x}_i^{j}$ to denote chunk $\boldsymbol{x}_i$ embedded with temporal index $j$:
\begin{equation}
\small
\begin{aligned}
\boldsymbol{x}^{j}_{i} &= \texttt{RoPE}(\boldsymbol{x}_{i}, j),\\
\boldsymbol{x}^{j}_{i}[k] &= \texttt{RoPE}(\boldsymbol{z}_{in+k}, jn+k),\ k\in[0,n).
\end{aligned}
\end{equation}
When no superscript is specified, we have $j\equiv i$.
Following Eq.~\ref{eq:as}, Temporal Re-index assigns the $S$ sink chunks to the sliding temporal window $[i-K,i-(K-S))$:
\begin{equation}
\label{eq:tr}
\small
\begin{aligned}
&\quad
p_\theta\left(
\boldsymbol{x}_i
\mid
\texttt{Cat}\left(
\boldsymbol{x}^{[i-K,i-(K-S))}_{[0,S)},
\boldsymbol{x}_{[i-(K-S),i)}
\right)
\right),
\\
&\text{where}\ i\in[K,\infty),\ \boldsymbol{x}^{[i-K,i-(K-S))}_{[0,S)}=\left\{\boldsymbol{x}^{i-K+l}_{l}\right\}_{l\in[0,S)}.
\end{aligned}
\end{equation}
Temporal Re-index restores sliding temporal indices while preserving low-drift content, but the sink's semantics are still fixed to the same early chunks $\boldsymbol{x}_{[0,S)}$.

\setlength{\cmidrulewidth}{\lightrulewidth}

\begin{table*}[!t]
\centering
\caption{
\textbf{1-minute AR video synthesis: comparison with training-free methods.}
We compare Rolling Sink with training-free horizon-extension methods on two 5s-trained bases, Self Forcing~\citep{huang2025self} and Causal Forcing~\citep{zhu2026causal}, using \texttt{VBench-Long}~\citep{huang2023vbench,huang2025vbench++,zheng2025vbench2}.
Rows are grouped by base model, each group holds the same base and inference setting.
Rolling Sink attains the best averaged rank in both groups.
The best and second-best scores within each group are in \textbf{bold} and \underline{underline}.
\emph{Forward-only} drops the reverse leg of Eq.~\ref{eq:roll}, cycling the bank forward only ($\texttt{Roll}(\boldsymbol{x}_{[0,K)})[l]=\boldsymbol{x}^{l}_{l\bmod K}$), so the sink repeats with period $K$.
}
\vspace{-.2cm}
\resizebox{\textwidth}{!}{%
\label{tab:1min}
\begin{tabular}{lccccccccccccccccc}
\toprule
\multirow{2}{*}{Method} 
& \multicolumn{16}{c}{\texttt{VBench-Long} Dimensions $\uparrow$} 
& Avg \\
\cmidrule(lr){2-17}
& SubCon & BgCon & AesQl & ImgQl & ObjCls & MulObj & Col & SpaRel & Sce & TmpSty & OvrCon & HumAct & TmpFli & MoSmo & DyDg & AppSty & Rank$\downarrow$ \\
\midrule
\rowcolor{GroupGray}
\multicolumn{18}{c}{\emph{Base AR model: Self Forcing}} \\
Self Forcing & 96.79 & 96.53 & 59.16 & 69.80 & 86.80 & 36.39 & 64.33 & 71.21 & 10.79 & 22.20 & 19.91 & 68.86 & 97.63 & 98.14 & 48.57 & \textbf{20.99} & 5.06 \\
Deep Forcing & 97.83 & 95.92 & 60.20 & 69.48 & \textbf{100.00} & 58.34 & 80.35 & 99.07 & 18.10 & 24.34 & 22.74 & \textbf{88.57} & 97.57 & 98.33 & 75.00 & 19.33 & 3.75 \\
$\infty$-RoPE & 98.17 & 96.23 & 61.70 & \underline{72.09} & \textbf{100.00} & 58.20 & 80.93 & 98.11 & 17.46 & 23.41 & 20.97 & \underline{83.71} & 97.62 & 97.81 & \textbf{76.07} & \underline{20.21} & 3.62 \\
MemRoPE & 98.41 & 96.20 & 58.49 & \textbf{72.21} & 99.69 & \textbf{81.19} & \textbf{92.06} & 98.56 & \textbf{25.16} & 23.93 & 21.82 & 75.88 & 97.94 & \underline{98.56} & 62.87 & 19.50 & 3.25 \\
\rowcolor{OursTint}
Rolling Sink & \textbf{98.58} & \textbf{96.94} & \textbf{63.08} & 69.68 & \textbf{100.00} & \underline{69.98} & 80.23 & \textbf{100.00} & \underline{21.59} & \textbf{25.03} & \underline{23.16} & 78.00 & \textbf{98.16} & \textbf{98.65} & 74.69 & 18.91 & \textbf{2.19} \\
\rowcolor{OursTint}
(Forward Only) & \underline{98.45} & \underline{96.75} & \underline{62.16} & 68.95 & \textbf{100.00} & 68.64 & \underline{81.73} & \underline{99.99} & 19.24 & \underline{24.95} & \textbf{23.35} & 75.33 & \underline{98.11} & 98.48 & \underline{75.36} & 18.70 & \underline{2.75} \\
\midrule
\rowcolor{GroupGray}
\multicolumn{18}{c}{\emph{Base AR model: Causal Forcing}} \\
Causal Forcing & 95.65 & \underline{96.11} & 52.73 & 62.86 & 69.02 & 32.23 & 74.53 & 65.91 & 13.02 & 18.39 & 17.73 & 50.57 & \textbf{96.30} & 97.17 & 54.64 & \textbf{21.13} & 4.25 \\
Deep Forcing & \underline{97.49} & 95.18 & 61.05 & 68.99 & 89.50 & 35.91 & \underline{84.33} & 92.09 & 16.83 & 22.19 & 20.39 & 75.14 & 95.38 & \underline{97.48} & \underline{93.57} & 20.34 & 3.25 \\
MemRoPE & 97.00 & 94.80 & 60.83 & \textbf{71.26} & 94.89 & \underline{40.92} & 79.13 & \underline{96.98} & 16.01 & \textbf{23.24} & 21.11 & 76.47 & 93.53 & 97.39 & 88.24 & \underline{20.55} & 3.00 \\
$\infty$-RoPE & 96.99 & 95.40 & \textbf{62.68} & 70.16 & \underline{96.12} & 39.88 & 80.14 & 95.77 & \underline{21.90} & \underline{23.16} & \underline{21.61} & \underline{82.29} & 95.44 & 95.46 & \textbf{98.93} & 20.02 & \underline{2.69} \\
\rowcolor{OursTint}
Rolling Sink & \textbf{98.16} & \textbf{96.38} & \underline{62.42} & \underline{70.45} & \textbf{98.66} & \textbf{74.96} & \textbf{89.02} & \textbf{98.52} & \textbf{22.86} & 21.86 & \textbf{23.30} & \textbf{84.29} & \underline{96.24} & \textbf{98.04} & 78.57 & 20.00 & \textbf{1.81} \\
\bottomrule
\end{tabular}%
}
\end{table*}
\setlength{\cmidrulewidth}{\lightrulewidth}

\begin{table*}[t]
\centering
\caption{
\textbf{5-minute AR video synthesis: comparison with training-free methods.}
This setting is harder than Tab.~\ref{tab:1min}, since drift has longer to accumulate.
Rolling Sink again attains the best averaged rank on both the Self Forcing~\citep{huang2025self} and Causal Forcing~\citep{zhu2026causal} bases.
Following Tab.~\ref{tab:1min}, rows are grouped by base model.
Best and second-best within each group are in \textbf{bold} and \underline{underline}.
}
\vspace{-.2cm}
\resizebox{\textwidth}{!}{%
\label{tab:5min}
\begin{tabular}{lccccccccccccccccc}
\toprule
\multirow{2}{*}{Method} 
& \multicolumn{16}{c}{\texttt{VBench-Long} Dimensions $\uparrow$} 
& Avg \\
\cmidrule(lr){2-17}
& SubCon & BgCon & AesQl & ImgQl & ObjCls & MulObj & Col & SpaRel & Sce & TmpSty & OvrCon & HumAct & TmpFli & MoSmo & DyDg & AppSty & Rank$\downarrow$ \\
\midrule
\rowcolor{GroupGray}
\multicolumn{18}{c}{\emph{Base AR model: Self Forcing}} \\
Self Forcing & 94.24 & \underline{96.10} & 42.89 & 57.01 & 33.39 & 14.27 & 61.05 & 35.70 & 4.30 & 11.32 & 11.39 & 27.10 & \underline{98.20} & \textbf{98.65} & 24.19 & \textbf{20.53} & 4.88 \\
MemRoPE & 96.69 & 94.94 & 57.94 & \underline{71.76} & 99.64 & 62.66 & 73.61 & 95.46 & 23.66 & 21.35 & 21.82 & 80.65 & 97.79 & 98.47 & 65.32 & 19.47 & 4.25 \\
$\infty$-RoPE & 97.14 & 94.43 & 59.00 & \textbf{71.82} & 89.15 & 49.09 & \underline{82.41} & 95.21 & \underline{27.24} & 21.97 & 19.99 & \textbf{90.32} & 97.88 & 97.76 & \underline{70.56} & \underline{19.72} & 3.50 \\
Deep Forcing & 96.76 & 94.59 & 60.54 & 69.29 & 99.90 & 60.52 & \textbf{86.65} & 98.97 & \textbf{33.33} & 23.72 & 22.16 & 83.55 & 97.88 & 97.67 & \textbf{78.23} & 19.57 & 3.25 \\
\rowcolor{OursTint}
Rolling Sink & \textbf{98.04} & \textbf{96.29} & \textbf{62.96} & 69.87 & \textbf{100.00} & \textbf{72.84} & 78.83 & \textbf{100.00} & 26.16 & \textbf{25.33} & \underline{23.10} & \underline{87.10} & \textbf{98.32} & \underline{98.59} & 64.11 & 18.91 & \textbf{2.12} \\
\rowcolor{OursTint}
(Forward Only) & \underline{97.82} & 96.03 & \underline{62.47} & 69.23 & \textbf{100.00} & \underline{71.57} & 81.13 & \textbf{100.00} & 25.47 & \underline{24.98} & \textbf{23.30} & 86.57 & 98.16 & 98.53 & 65.03 & 18.75 & \underline{2.81} \\
\midrule
\rowcolor{GroupGray}
\multicolumn{18}{c}{\emph{Base AR model: Causal Forcing}} \\
Causal Forcing & 92.35 & \underline{94.94} & 40.30 & 55.12 & 29.11 & 8.33 & 75.05 & 21.40 & 4.66 & 11.33 & 11.10 & 38.39 & \textbf{96.97} & 96.70 & 31.45 & \textbf{20.80} & 4.19 \\
Deep Forcing & 94.74 & 92.81 & 58.37 & 68.14 & 85.54 & 29.84 & 79.13 & 85.93 & 16.13 & 21.84 & 20.31 & \textbf{86.13} & 94.90 & \underline{96.95} & 89.52 & \underline{20.69} & 3.25 \\
MemRoPE & 94.16 & 92.49 & 59.61 & 68.85 & \underline{91.83} & \underline{38.10} & 71.88 & 88.15 & 15.05 & \textbf{25.08} & 22.23 & \underline{80.00} & 93.50 & 96.75 & \underline{93.95} & 20.46 & 3.12 \\
$\infty$-RoPE & \underline{95.19} & 93.32 & \textbf{62.86} & \underline{70.15} & \textbf{95.02} & 37.36 & \underline{81.51} & \underline{92.03} & \textbf{26.16} & \underline{24.98} & \underline{23.21} & \underline{80.00} & 95.33 & 95.41 & \textbf{99.60} & 20.39 & \underline{2.25} \\
\rowcolor{OursTint}
Rolling Sink & \textbf{97.09} & \textbf{95.57} & \underline{61.97} & \textbf{70.32} & 91.79 & \textbf{72.80} & \textbf{88.29} & \textbf{97.84} & \underline{17.56} & 21.55 & \textbf{23.27} & 78.39 & \underline{96.26} & \textbf{98.16} & 84.68 & 20.12 & \textbf{2.12} \\
\bottomrule
\end{tabular}%
}
\vspace{-.2cm}
\end{table*}
\begin{figure*}[!t]
    \centering
    \includegraphics[width=\linewidth]{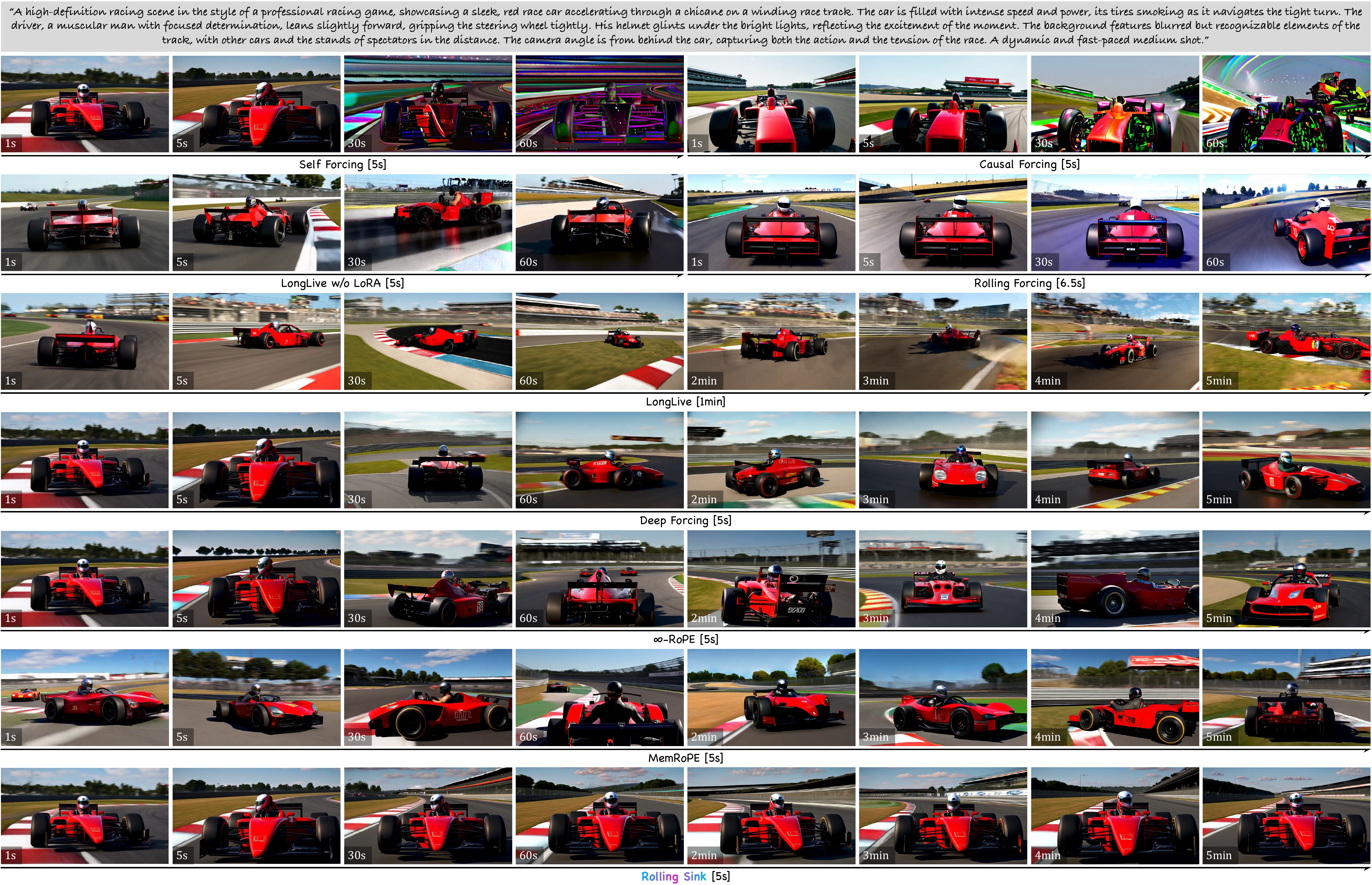}
\vspace{-.5cm}
\caption{
\textbf{Qualitative comparison on long-horizon AR video synthesis.}
We compare Rolling Sink with training-based and training-free horizon-extension methods; each method's training duration is in square brackets.
For Rolling Sink and the training-free baselines, the base model is Self Forcing~\citep{huang2025self}.
Rolled out beyond their training horizons, the baselines accumulate drift: oversaturated colors, distorted structure, subject change, and fading motion.
Rolling Sink holds subject, structure, and motion more stable across the rollout while using only a 5s base, including against baselines trained on minute-long clips.
}
    \label{fig:qc}
\vspace{-.3cm}
\end{figure*}

\subsubsection{Rolling Semantics: Evolving Semantic Content}
\label{sec:roll_sem}

Temporal Re-index still pins the sink to the same early chunks $\boldsymbol{x}_{[0,S)}$, so its semantics stay fixed while the rollout moves on.
Rolling Semantics lets that content evolve.
It treats the finite bank of $K$ low-drift chunks $\boldsymbol{x}_{[0,K)}$, generated within the training horizon, as ordered samples along a semantic axis.
At each step it reads a sliding window of $S$ consecutive positions from this axis. The sink content then shifts by one position per chunk instead of freezing, mirroring the within-horizon rollout where the semantic window moves chunk by chunk.
The conditional distribution becomes:
\begin{equation}
\label{eq:rs}
\small
\begin{aligned}
&\quad
p_\theta\left(
\boldsymbol{x}_i
\mid
\texttt{Cat}\left(
\texttt{Roll}(\boldsymbol{x}_{[0,K)})_{[i-K,i-(K-S))},
\boldsymbol{x}_{[i-(K-S),i)}
\right)
\right),\\
&\text{where}\ i\in[K,\infty),\
\texttt{Roll}(\boldsymbol{x}_{[0,K)})_{[a,b)}
=
\left\{\texttt{Roll}(\boldsymbol{x}_{[0,K)})[l]\right\}_{l\in[a,b)}.
\end{aligned}
\end{equation}

The bank is finite, so a long rollout eventually revisits its chunks: naive cycling does this by jumping from the last chunk $\boldsymbol{x}_{K-1}$ back to $\boldsymbol{x}_0$ every $K$ steps.
Rolling Semantics instead traverses the bank forward then backward, reversing the frame order within a chunk on the backward pass, so successive read-outs stay adjacent and no periodic jump appears:
\begin{equation}
\label{eq:roll}
\small
\texttt{Roll}(\boldsymbol{x}_{[0,K)})[l]=
\left\{
\begin{aligned}
&\boldsymbol{x}^{l}_{l\bmod K}
&\text{if } \left\lfloor\frac{l}{K}\right\rfloor \bmod 2=0,\\
&\boldsymbol{\tilde{x}}^{l}_{(K-1)-(l\bmod K)}
&\text{if } \left\lfloor\frac{l}{K}\right\rfloor \bmod 2=1,
\end{aligned}
\right.
\end{equation}
where $l\in[0,\infty)$ and $\boldsymbol{\tilde{x}}_i^j$ is the frame-reversed form of $\boldsymbol{x}_i^j$, with $\boldsymbol{\tilde{x}}_i^j[k]=\texttt{RoPE}(\boldsymbol{z}_{in+(n-1-k)}, jn+k)$ for $k\in[0,n)$.

Retaining one recent chunk beside the rolling sink anchors each prediction to the most recent context, keeping local motion smooth and the output moving forward rather than replaying the bank (App.~\ref{supp:sec:rep} and~\ref{supp:sec:periodicity}).
The backward pass reorders frames only inside the sink chunks, which supply content and semantics rather than the visible motion.
The motion is carried by the recent chunk, which is not reversed, so the reversal does not play the visible dynamics backward.
Rolling Sink is the resulting policy: under the bounded budget it holds low-drift content, sliding temporal indices, and evolving semantics together.
This re-scheduling of within-horizon content narrows the cache-behavior mismatch between limited-horizon training and open-ended testing, but cannot fully close it (Sec.~\ref{sec:con}).
App.~\ref{supp:sec:algo} gives the full pseudo-code.

\section{Experiments}
\label{sec:exp}

\subsection{Experimental Setup}

\noindent\textbf{Implementation.}
Rolling Sink is training-free.
We apply it to two 5s-trained bases, Self Forcing~\citep{huang2025self} and Causal Forcing~\citep{zhu2026causal}, and reuse each base's sampling configuration: the same few-step denoising and the same bounded cache budget $K=6$.
We set $S/K=5/6$ (ablated in Sec.~\ref{sec:ablation}).

\noindent\textbf{Protocol.}
We evaluate two rollout lengths, 1 minute and 5 minutes, with \texttt{VBench-Long}~\citep{huang2023vbench,huang2025vbench++,zheng2025vbench2}, which scores long videos on 16 dimensions. We set its \texttt{clip\_length} to 2.0 for the 1-minute setting and 10.0 for the 5-minute one.
We report every dimension and the averaged rank, where a lower rank is better.
Averaged rank is computed within each same-base group.
We score a fixed subset of 10 prompts per dimension, drawn once from the original 70 to 100 and reused across all runs. App.~\ref{supp:sec:vbench} gives the full dimension names and this sampling protocol.
The training-free comparison holds the same base model and inference setting.
Qualitatively, we compare rollouts at matched timestamps for color shift, identity change, structural collapse, and motion loss.

\noindent\textbf{Baselines.}
We compare against two groups.
The training-free horizon-extension group includes Deep Forcing~\citep{yi2025deep}, $\infty$-RoPE~\citep{yesiltepe2025infinity}, and MemRoPE~\citep{kim2026memrope}.
The training-based group includes Self Forcing~\citep{huang2025self}, Causal Forcing~\citep{zhu2026causal}, Rolling Forcing~\citep{liu2025rolling}, and LongLive~\citep{yang2025longlive}.

\subsection{Comparison with Training-Free Methods}

Tab.~\ref{tab:1min} and Tab.~\ref{tab:5min} compare Rolling Sink with training-free horizon-extension methods.
Rolling Sink attains the best averaged rank in all four groups (2.19 and 1.81 at 1 minute on Self Forcing and Causal Forcing, 2.12 and 2.12 at 5 minutes), consistent across both lengths and bases.
This supports treating the horizon gap as cache-behavior matching rather than a single-axis fix.
The Forward-only rows isolate the schedule: forward-only cycling of the bank trails the full forward-reverse schedule on averaged rank, so the reverse leg helps beyond simple cycling.
Averaged rank aggregates 16 dimensions, so the lead is an aggregate one under the evaluated protocol, not a win on every dimension.

Per dimension, the gains concentrate on the consistency, quality, and temporal-stability axes: on the 1-minute Self Forcing base, Rolling Sink leads on subject and background consistency, aesthetic quality, spatial relationship, temporal style, temporal flickering, and motion smoothness.
It trades off elsewhere, with MemRoPE leading on imaging quality, multiple objects, color, and scene, Deep Forcing on human action, $\infty$-RoPE on dynamic degree, and the frozen base on appearance style by about two points, so the averaged rank is a balance rather than a uniform lead.

\subsection{Comparison with Training-Based Baselines}
\label{sec:training_baseline}

Tab.~\ref{tab:training_baseline} adds training-based baselines, several trained on longer clips (6.5s for Rolling Forcing, 1 minute for LongLive).
Rolling Sink on the 5s Self Forcing base reaches the best averaged rank at both lengths (2.38 and 2.06), ahead of LongLive trained on 1-minute clips (2.69 and 2.63), despite using no extra training and a shorter horizon.

\setlength{\cmidrulewidth}{\lightrulewidth}

\begin{table*}[t]
\centering
\caption{
\textbf{Comparison with training-based AR baselines.}
We compare Rolling Sink with training-based AR baselines at 1-minute and 5-minute rollouts on \texttt{VBench-Long}~\citep{huang2023vbench,huang2025vbench++,zheng2025vbench2}.
Superscripts give each method's training duration; the Rolling Sink subscript gives its base, Self Forcing (SF)~\citep{huang2025self} or Causal Forcing (CF)~\citep{zhu2026causal}.
Rolling Sink on the SF base reaches the best averaged rank at both lengths, even ahead of LongLive~\citep{yang2025longlive} trained on 1-minute clips, while using no extra training on a 5s base.
The CF variant ranks third, behind LongLive-1min.
The best and second-best scores are in \textbf{bold} and \underline{underline}.
}
\vspace{-.2cm}
\resizebox{\textwidth}{!}{%
\label{tab:training_baseline}
\begin{tabular}{lccccccccccccccccc}
\toprule
\multirow{2}{*}{Method} 
& \multicolumn{16}{c}{\texttt{VBench-Long} Dimensions $\uparrow$} 
& Avg \\
\cmidrule(lr){2-17}
& SubCon & BgCon & AesQl & ImgQl & ObjCls & MulObj & Col & SpaRel & Sce & TmpSty & OvrCon & HumAct & TmpFli & MoSmo & DyDg & AppSty & Rank$\downarrow$ \\
\midrule
\rowcolor{GroupGray}
\multicolumn{18}{c}{\emph{1-minute AR Video Synthesis}} \\
Causal Forcing$^{\text{5s}}$ & 95.65 & 96.11 & 52.73 & 62.86 & 69.02 & 32.23 & 74.53 & 65.91 & 13.02 & 18.39 & 17.73 & 50.57 & 96.30 & 97.17 & 54.64 & \textbf{21.13} & 6.25 \\
Self Forcing$^{\text{5s}}$ & 96.79 & \underline{96.53} & 59.16 & \underline{69.80} & 86.80 & 36.39 & 64.33 & 71.21 & 10.79 & 22.20 & 19.91 & 68.86 & 97.63 & 98.14 & 48.57 & \underline{20.99} & 4.81 \\
LongLive$^{\text{5s}}_{\text{w/o LoRA}}$ & 96.68 & 95.88 & 58.50 & 65.19 & 97.80 & 58.02 & 77.12 & 96.83 & \textbf{25.40} & 23.98 & 21.60 & \textbf{88.57} & 96.43 & 97.30 & 70.71 & 20.18 & 4.50 \\
Rolling Forcing$^{\text{6.5s}}$ & 97.95 & 96.22 & 60.68 & 68.09 & \textbf{100.00} & 62.25 & 84.97 & 99.58 & \underline{24.13} & \underline{24.12} & 20.49 & 74.00 & 97.33 & 97.87 & 32.86 & 20.23 & 3.88 \\
\rowcolor{OursTint}
Rolling Sink$^{\text{5s}}_{\text{CF}}$ & 98.16 & 96.38 & 62.42 & \textbf{70.45} & 98.66 & \underline{74.96} & \underline{89.02} & 98.52 & 22.86 & 21.86 & \textbf{23.30} & \underline{84.29} & 96.24 & 98.04 & \textbf{78.57} & 20.00 & 3.25 \\
LongLive$^{\text{1min}}$ & \underline{98.40} & 96.50 & \underline{62.56} & 69.47 & \textbf{100.00} & \textbf{88.64} & \textbf{98.66} & \textbf{100.00} & 15.87 & 23.29 & \underline{23.21} & 80.57 & \textbf{98.18} & \underline{98.48} & 55.00 & 18.77 & \underline{2.69} \\
\rowcolor{OursTint}
Rolling Sink$^{\text{5s}}_{\text{SF}}$ & \textbf{98.58} & \textbf{96.94} & \textbf{63.08} & 69.68 & \textbf{100.00} & 69.98 & 80.23 & \textbf{100.00} & 21.59 & \textbf{25.03} & 23.16 & 78.00 & \underline{98.16} & \textbf{98.65} & \underline{74.69} & 18.91 & \textbf{2.38} \\
\midrule
\rowcolor{GroupGray}
\multicolumn{18}{c}{\emph{5-minute AR Video Synthesis}} \\
Causal Forcing$^{\text{5s}}$ & 92.35 & 94.94 & 40.30 & 55.12 & 29.11 & 8.33 & 75.05 & 21.40 & 4.66 & 11.33 & 11.10 & 38.39 & 96.97 & 96.70 & 31.45 & \underline{20.80} & 6.06 \\
Self Forcing$^{\text{5s}}$ & 94.24 & \underline{96.10} & 42.89 & 57.01 & 33.39 & 14.27 & 61.05 & 35.70 & 4.30 & 11.32 & 11.39 & 27.10 & 98.20 & \textbf{98.65} & 24.19 & 20.53 & 5.31 \\
LongLive$^{\text{5s}}_{\text{w/o LoRA}}$ & 93.93 & 94.27 & 57.18 & 64.31 & 96.65 & 69.98 & 73.02 & 96.97 & 20.79 & 24.35 & 21.50 & \textbf{95.48} & 96.87 & 97.66 & \underline{73.79} & \textbf{20.86} & 4.31 \\
Rolling Forcing$^{\text{6.5s}}$ & 96.72 & 95.09 & 58.09 & 67.70 & \textbf{100.00} & 71.73 & 77.63 & 98.57 & \textbf{34.41} & 23.84 & 20.20 & 71.94 & 97.58 & 98.04 & 31.05 & 20.21 & 3.94 \\
\rowcolor{OursTint}
Rolling Sink$^{\text{5s}}_{\text{CF}}$ & \underline{97.09} & 95.57 & 61.97 & \textbf{70.32} & 91.79 & 72.80 & \textbf{88.29} & 97.84 & 17.56 & 21.55 & \textbf{23.27} & 78.39 & 96.26 & 98.16 & \textbf{84.68} & 20.12 & 3.44 \\
LongLive$^{\text{1min}}$ & 96.91 & 96.01 & \textbf{63.70} & 69.78 & \textbf{100.00} & \textbf{76.90} & \underline{82.80} & \textbf{100.00} & 20.07 & \underline{24.57} & 22.66 & 86.13 & \textbf{98.35} & 98.46 & 59.68 & 18.54 & \underline{2.63} \\
\rowcolor{OursTint}
Rolling Sink$^{\text{5s}}_{\text{SF}}$ & \textbf{98.04} & \textbf{96.29} & \underline{62.96} & \underline{69.87} & \textbf{100.00} & \underline{72.84} & 78.83 & \textbf{100.00} & \underline{26.16} & \textbf{25.33} & \underline{23.10} & \underline{87.10} & \underline{98.32} & \underline{98.59} & 64.11 & 18.91 & \textbf{2.06} \\
\bottomrule
\end{tabular}%
}
\vspace{-.2cm}
\end{table*}
\begin{figure*}[!t]
    \centering
    \includegraphics[width=\linewidth]{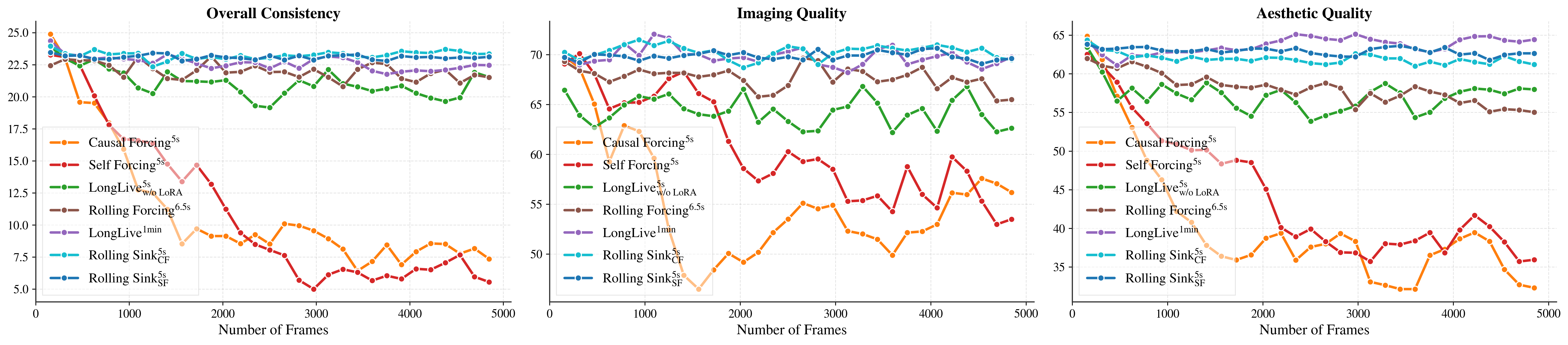}
\vspace{-.5cm}
\caption{
\textbf{Drift over the 5-minute rollout.}
We track three \texttt{VBench-Long}~\citep{huang2023vbench,huang2025vbench++,zheng2025vbench2} dimensions, Overall Consistency, Imaging Quality, and Aesthetic Quality, along the generation trajectory.
Baselines degrade or fluctuate as frames accumulate, reflecting drift.
Rolling Sink stays higher and flatter across the rollout, keeping quality high while slowing drift.
}    \label{fig:drift}
\vspace{-.3cm}
\end{figure*}

\subsection{Cache-Behavior Ablation}
\label{sec:ablation}
\begin{wrapfigure}{r}{0.5\textwidth}
    \centering
    \vspace{-.3cm}
    \includegraphics[width=0.5\textwidth]{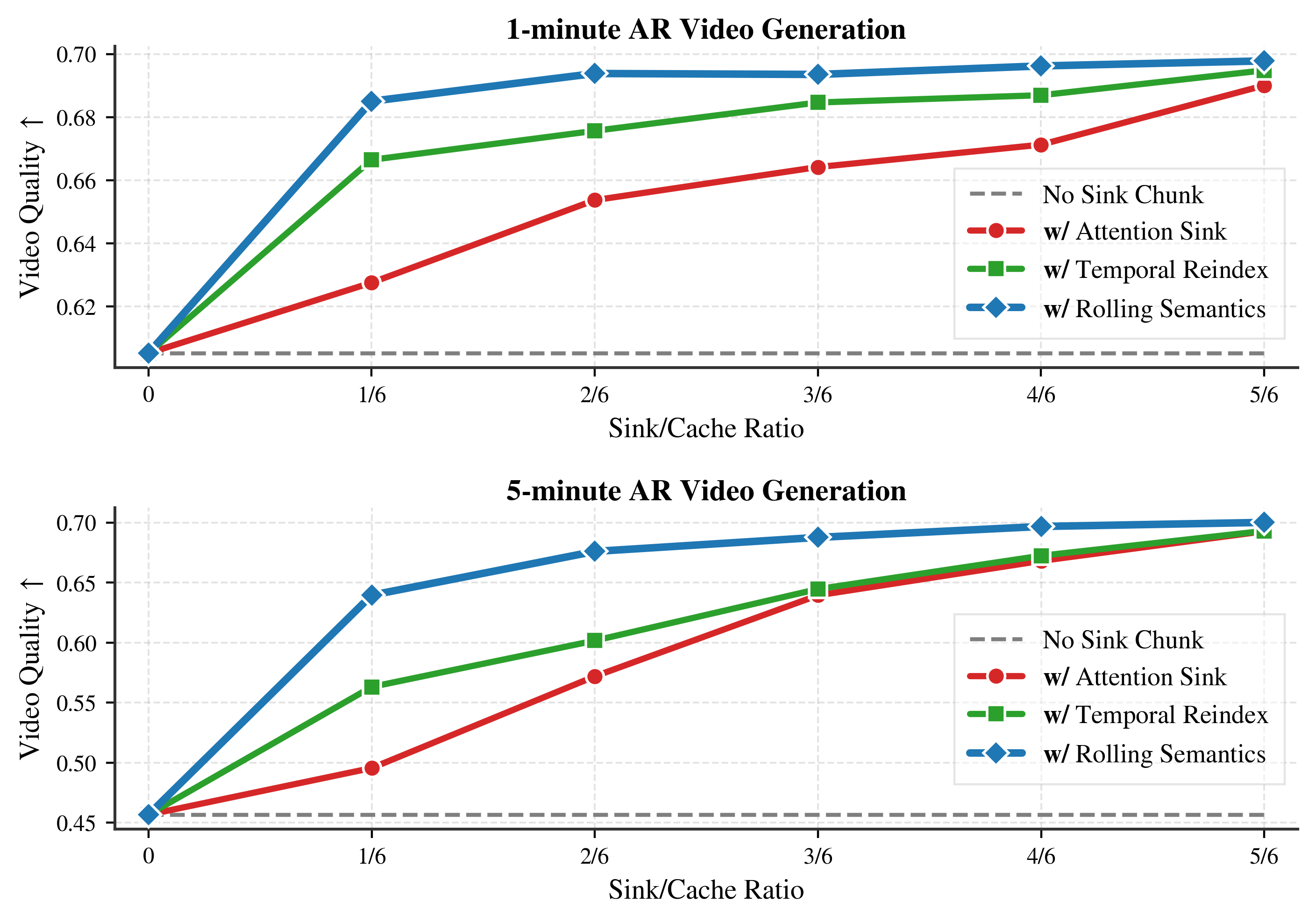}
\vspace{-.57cm}
\caption{
\textbf{Cache-behavior ablation.}
\emph{Video Quality} (mean over all \texttt{VBench-Long} dimensions) rises as each cache behavior is added, best at $S/K=5/6$.
}
    \label{fig:abl}
\vspace{-.6cm}
\end{wrapfigure}

Fig.~\ref{fig:abl} scores each stage of the derivation (Sec.~\ref{sec:method_rs}) on \texttt{VBench-Long}~\citep{huang2023vbench,zheng2025vbench2,huang2025vbench++}, plotting \emph{Video Quality}, the mean over all 16 dimensions (the Avg Score of Tab.~\ref{supp:tab:abl_1min} and~\ref{supp:tab:abl_5min}).
Adding each cache behavior raises this score at both 1 and 5 minutes.
Attention Sink provides the largest single gain by restoring low-drift content. Temporal Re-index and Rolling Semantics add smaller further gains, and Rolling Semantics reaches the best final score at the default $S/K=5/6$.
The full per-dimension numbers under every sink ratio are in Tab.~\ref{supp:tab:abl_1min} and~\ref{supp:tab:abl_5min}.
We use $S/K=5/6$ as the default, its best-performing setting\footnote{$S/K=5/6$ keeps five rolling sink chunks and one recent chunk under the shared budget $K=6$. We sweep $S/K$ within this budget, since other budgets would require different bases.}.

\subsection{Qualitative Comparison and Drift Analysis}

Fig.~\ref{fig:qc} shows rollouts against both baseline groups (base model Self Forcing).
$\infty$-RoPE extends positions but does not restore low-drift content, and Deep Forcing and MemRoPE retain a sink or memory prefix with fixed semantics, so all three still show color shift, structural change, or subject drift.
Training-based baselines also drift once rolled far past their trained length.
Rolling Sink holds subject, structure, and motion more stable across the rollout while using only a 5s base.

Fig.~\ref{fig:drift} tracks three \texttt{VBench-Long} dimensions along the 5-minute trajectory: baselines degrade or fluctuate as frames accumulate, whereas Rolling Sink stays higher and flatter. Keeping the cache near its within-horizon behavior both raises quality and slows drift.

\subsection{Inference Efficiency}

Rolling Sink keeps the base's cache size and denoising steps, so its throughput tracks the base: 5.35 vs.~5.38 FPS on an RTX 4090 at the 1-minute setting, a 0.56\% change (App.~\ref{supp:sec:fps}).
The bounded cache also holds GPU memory constant, since $K$ does not grow with rollout length.

\section{Conclusion and Limitations}
\label{sec:con}

We cast long-horizon AR drift as a domain shift on the cache and propose Rolling Sink, a training-free bounded-cache policy that matches the within-horizon cache along content, position, and semantics.
It reuses attention sink and temporal re-indexing and adds the Rolling Semantics schedule.
It attains the best averaged \texttt{VBench-Long} rank at 1- and 5-minute rollouts on two 5s bases at about 0.56\% throughput change, and sustains a 30-minute qualitative rollout.

\noindent\textbf{Limitations.}
Under limited-horizon training, the gap can only be reduced, not closed: the base never learned the long-range content that open-ended testing needs.
Rolling Sink matches content and position near-exactly (reusing low-drift chunks, re-indexed to the exact window) but only approximates semantics: the finite bank cannot synthesize new long-range content, it reads out the same chunks periodically, and its reverse legs run the sink's semantics backward (a partial time-reversal).
This residual needs more training or data, not a better cache.
Matching all three factors (Tab.~\ref{tab:method}) is the closest a training-free policy can come.
We analyze return-to-start repetition in App.~\ref{supp:sec:rep} and periodic replay and progression in App.~\ref{supp:sec:periodicity}.
Several directions remain open, such as behavior on individual irreversible events (an object falling, liquid pouring, something breaking).

\section{AI Tools in Paper Writing}

We used generative AI tools solely for polishing grammar and sentence structures, to improve readability, clarity, and fluency.
They made no contribution to the original research content.
All the scientific and technical content of this paper was written entirely by humans.
We have reviewed all AI-assisted work and we take responsibility for all the final content of this work.

\bibliography{main}

@misc{sora,
  title={Sora 2 is here},
  author={OpenAI},
  year={2025},
  howpublished = {\url{https://openai.com/index/sora-2/}}
}

@misc{gen4,
  title={Introducing Runway Gen-4.5: A new frontier for video generation},
  author={Runway},
  year={2025},
  howpublished = {\url{https://runwayml.com/research/introducing-runway-gen-4.5}}
}

@misc{veo3,
  title={Introducing Veo 3, our video generation model with expanded creative controls – including native audio and extended videos},
  author={Google},
  year={2025},
  howpublished = {\url{https://deepmind.google/models/veo/}}
}

@misc{kling2.6,
  title={Kling VIDEO 2.6 – Kling's First ``Native Audio'' Model Official Launched!},
  author={Kling},
  year={2025},
  howpublished = {\url{https://app.klingai.com/global/release-notes/c605hp1tzd}}
}

@article{xiang2026pathwise,
  title={Pathwise test-time correction for autoregressive long video generation},
  author={Xiang, Xunzhi and Duan, Zixuan and Zhang, Guiyu and Zhang, Haiyu and Gao, Zhe and Wu, Junta and Zhang, Shaofeng and Wang, Tengfei and Fan, Qi and Guo, Chunchao},
  journal={arXiv preprint arXiv:2602.05871},
  year={2026}
}

@article{wan2025,
      title={Wan: Open and Advanced Large-Scale Video Generative Models}, 
      author={Team Wan and Ang Wang and Baole Ai and Bin Wen and Chaojie Mao and Chen-Wei Xie and Di Chen and Feiwu Yu and Haiming Zhao and Jianxiao Yang and Jianyuan Zeng and Jiayu Wang and Jingfeng Zhang and Jingren Zhou and Jinkai Wang and Jixuan Chen and Kai Zhu and Kang Zhao and Keyu Yan and Lianghua Huang and Mengyang Feng and Ningyi Zhang and Pandeng Li and Pingyu Wu and Ruihang Chu and Ruili Feng and Shiwei Zhang and Siyang Sun and Tao Fang and Tianxing Wang and Tianyi Gui and Tingyu Weng and Tong Shen and Wei Lin and Wei Wang and Wei Wang and Wenmeng Zhou and Wente Wang and Wenting Shen and Wenyuan Yu and Xianzhong Shi and Xiaoming Huang and Xin Xu and Yan Kou and Yangyu Lv and Yifei Li and Yijing Liu and Yiming Wang and Yingya Zhang and Yitong Huang and Yong Li and You Wu and Yu Liu and Yulin Pan and Yun Zheng and Yuntao Hong and Yupeng Shi and Yutong Feng and Zeyinzi Jiang and Zhen Han and Zhi-Fan Wu and Ziyu Liu},
      journal = {arXiv preprint arXiv:2503.20314},
      year={2025}
}

@inproceedings{peebles2023scalable,
  title={Scalable diffusion models with transformers},
  author={Peebles, William and Xie, Saining},
  booktitle={Proceedings of the IEEE/CVF international conference on computer vision},
  pages={4195--4205},
  year={2023}
}

@article{huang2025self,
  title={Self Forcing: Bridging the Train-Test Gap in Autoregressive Video Diffusion},
  author={Huang, Xun and Li, Zhengqi and He, Guande and Zhou, Mingyuan and Shechtman, Eli},
  journal={arXiv preprint arXiv:2506.08009},
  year={2025}
}

@article{yang2025longlive,
  title={Longlive: Real-time interactive long video generation},
  author={Yang, Shuai and Huang, Wei and Chu, Ruihang and Xiao, Yicheng and Zhao, Yuyang and Wang, Xianbang and Li, Muyang and Xie, Enze and Chen, Yingcong and Lu, Yao and others},
  journal={arXiv preprint arXiv:2509.22622},
  year={2025}
}

@article{liu2025rolling,
  title={Rolling forcing: Autoregressive long video diffusion in real time},
  author={Liu, Kunhao and Hu, Wenbo and Xu, Jiale and Shan, Ying and Lu, Shijian},
  journal={arXiv preprint arXiv:2509.25161},
  year={2025}
}

@article{cui2025self,
  title={Self-Forcing++: Towards Minute-Scale High-Quality Video Generation},
  author={Cui, Justin and Wu, Jie and Li, Ming and Yang, Tao and Li, Xiaojie and Wang, Rui and Bai, Andrew and Ban, Yuanhao and Hsieh, Cho-Jui},
  journal={arXiv preprint arXiv:2510.02283},
  year={2025}
}

@article{yi2025deep,
  title={Deep Forcing: Training-Free Long Video Generation with Deep Sink and Participative Compression},
  author={Yi, Jung and Jang, Wooseok and Cho, Paul Hyunbin and Nam, Jisu and Yoon, Heeji and Kim, Seungryong},
  journal={arXiv preprint arXiv:2512.05081},
  year={2025}
}

@article{xiao2023efficient,
  title={Efficient streaming language models with attention sinks},
  author={Xiao, Guangxuan and Tian, Yuandong and Chen, Beidi and Han, Song and Lewis, Mike},
  journal={arXiv preprint arXiv:2309.17453},
  year={2023}
}

@article{gao2024ca2,
  title={Ca2-vdm: Efficient autoregressive video diffusion model with causal generation and cache sharing},
  author={Gao, Kaifeng and Shi, Jiaxin and Zhang, Hanwang and Wang, Chunping and Xiao, Jun and Chen, Long},
  journal={arXiv preprint arXiv:2411.16375},
  year={2024}
}

@article{hu2024acdit,
  title={Acdit: Interpolating autoregressive conditional modeling and diffusion transformer},
  author={Hu, Jinyi and Hu, Shengding and Song, Yuxuan and Huang, Yufei and Wang, Mingxuan and Zhou, Hao and Liu, Zhiyuan and Ma, Wei-Ying and Sun, Maosong},
  journal={arXiv preprint arXiv:2412.07720},
  year={2024}
}

@article{jin2024pyramidal,
  title={Pyramidal flow matching for efficient video generative modeling},
  author={Jin, Yang and Sun, Zhicheng and Li, Ningyuan and Xu, Kun and Jiang, Hao and Zhuang, Nan and Huang, Quzhe and Song, Yang and Mu, Yadong and Lin, Zhouchen},
  journal={arXiv preprint arXiv:2410.05954},
  year={2024}
}

@article{zhang2025test,
  title={Test-time training done right},
  author={Zhang, Tianyuan and Bi, Sai and Hong, Yicong and Zhang, Kai and Luan, Fujun and Yang, Songlin and Sunkavalli, Kalyan and Freeman, William T and Tan, Hao},
  journal={arXiv preprint arXiv:2505.23884},
  year={2025}
}

@article{chen2024diffusion,
  title={Diffusion forcing: Next-token prediction meets full-sequence diffusion},
  author={Chen, Boyuan and Mart{\'\i} Mons{\'o}, Diego and Du, Yilun and Simchowitz, Max and Tedrake, Russ and Sitzmann, Vincent},
  journal={Advances in Neural Information Processing Systems},
  volume={37},
  pages={24081--24125},
  year={2024}
}

@inproceedings{yin2025slow,
  title={From slow bidirectional to fast autoregressive video diffusion models},
  author={Yin, Tianwei and Zhang, Qiang and Zhang, Richard and Freeman, William T and Durand, Fredo and Shechtman, Eli and Huang, Xun},
  booktitle={Proceedings of the Computer Vision and Pattern Recognition Conference},
  pages={22963--22974},
  year={2025}
}

@article{chen2025skyreels,
  title={Skyreels-v2: Infinite-length film generative model},
  author={Chen, Guibin and Lin, Dixuan and Yang, Jiangping and Lin, Chunze and Zhu, Junchen and Fan, Mingyuan and Zhang, Hao and Chen, Sheng and Chen, Zheng and Ma, Chengcheng and others},
  journal={arXiv preprint arXiv:2504.13074},
  year={2025}
}

@article{teng2025magi,
  title={MAGI-1: Autoregressive Video Generation at Scale},
  author={Teng, Hansi and Jia, Hongyu and Sun, Lei and Li, Lingzhi and Li, Maolin and Tang, Mingqiu and Han, Shuai and Zhang, Tianning and Zhang, WQ and Luo, Weifeng and others},
  journal={arXiv preprint arXiv:2505.13211},
  year={2025}
}

@article{song2025history,
  title={History-guided video diffusion},
  author={Song, Kiwhan and Chen, Boyuan and Simchowitz, Max and Du, Yilun and Tedrake, Russ and Sitzmann, Vincent},
  journal={arXiv preprint arXiv:2502.06764},
  year={2025}
}

@article{yin2024improved,
  title={Improved distribution matching distillation for fast image synthesis},
  author={Yin, Tianwei and Gharbi, Micha{\"e}l and Park, Taesung and Zhang, Richard and Shechtman, Eli and Durand, Fredo and Freeman, Bill},
  journal={Advances in neural information processing systems},
  volume={37},
  pages={47455--47487},
  year={2024}
}

@inproceedings{yin2024one,
  title={One-step diffusion with distribution matching distillation},
  author={Yin, Tianwei and Gharbi, Micha{\"e}l and Zhang, Richard and Shechtman, Eli and Durand, Fredo and Freeman, William T and Park, Taesung},
  booktitle={Proceedings of the IEEE/CVF conference on computer vision and pattern recognition},
  pages={6613--6623},
  year={2024}
}

@inproceedings{zhang2025frame,
  title={Frame context packing and drift prevention in next-frame-prediction video diffusion models},
  author={Zhang, Lvmin and Cai, Shengqu and Li, Muyang and Wetzstein, Gordon and Agrawala, Maneesh},
  booktitle={The Thirty-ninth Annual Conference on Neural Information Processing Systems},
  year={2025}
}

@article{huang2025vbench++,
    title={{VBench++}: Comprehensive and Versatile Benchmark Suite for Video Generative Models},
    author={Huang, Ziqi and Zhang, Fan and Xu, Xiaojie and He, Yinan and Yu, Jiashuo and Dong, Ziyue and Ma, Qianli and Chanpaisit, Nattapol and Si, Chenyang and Jiang, Yuming and Wang, Yaohui and Chen, Xinyuan and Chen, Ying-Cong and Wang, Limin and Lin, Dahua and Qiao, Yu and Liu, Ziwei},
    journal={IEEE Transactions on Pattern Analysis and Machine Intelligence}, 
    year={2025},
    doi={10.1109/TPAMI.2025.3633890}
}

@InProceedings{huang2023vbench,
    title={{VBench}: Comprehensive Benchmark Suite for Video Generative Models},
    author={Huang, Ziqi and He, Yinan and Yu, Jiashuo and Zhang, Fan and Si, Chenyang and Jiang, Yuming and Zhang, Yuanhan and Wu, Tianxing and Jin, Qingyang and Chanpaisit, Nattapol and Wang, Yaohui and Chen, Xinyuan and Wang, Limin and Lin, Dahua and Qiao, Yu and Liu, Ziwei},
    booktitle={Proceedings of the IEEE/CVF Conference on Computer Vision and Pattern Recognition},
    year={2024}
}

@article{zheng2025vbench2,
    title={{VBench-2.0}: Advancing Video Generation Benchmark Suite for Intrinsic Faithfulness},
    author={Zheng, Dian and Huang, Ziqi and Liu, Hongbo and Zou, Kai and He, Yinan and Zhang, Fan and Zhang, Yuanhan and He, Jingwen and Zheng, Wei-Shi and Qiao, Yu and Liu, Ziwei},
    journal={arXiv preprint arXiv:2503.21755},
    year={2025}
}

@article{blattmann2023stable,
  title={Stable video diffusion: Scaling latent video diffusion models to large datasets},
  author={Blattmann, Andreas and Dockhorn, Tim and Kulal, Sumith and Mendelevitch, Daniel and Kilian, Maciej and Lorenz, Dominik and Levi, Yam and English, Zion and Voleti, Vikram and Letts, Adam and others},
  journal={arXiv preprint arXiv:2311.15127},
  year={2023}
}

@article{kong2024hunyuanvideo,
  title={Hunyuanvideo: A systematic framework for large video generative models},
  author={Kong, Weijie and Tian, Qi and Zhang, Zijian and Min, Rox and Dai, Zuozhuo and Zhou, Jin and Xiong, Jiangfeng and Li, Xin and Wu, Bo and Zhang, Jianwei and others},
  journal={arXiv preprint arXiv:2412.03603},
  year={2024}
}

@article{yang2024cogvideox,
  title={Cogvideox: Text-to-video diffusion models with an expert transformer},
  author={Yang, Zhuoyi and Teng, Jiayan and Zheng, Wendi and Ding, Ming and Huang, Shiyu and Xu, Jiazheng and Yang, Yuanming and Hong, Wenyi and Zhang, Xiaohan and Feng, Guanyu and others},
  journal={arXiv preprint arXiv:2408.06072},
  year={2024}
}

@article{ren2025next,
  title={Next block prediction: Video generation via semi-autoregressive modeling},
  author={Ren, Shuhuai and Ma, Shuming and Sun, Xu and Wei, Furu},
  journal={arXiv preprint arXiv:2502.07737},
  year={2025}
}

@article{he2024lotus,
  title={Lotus: Diffusion-based visual foundation model for high-quality dense prediction},
  author={He, Jing and Li, Haodong and Yin, Wei and Liang, Yixun and Li, Leheng and Zhou, Kaiqiang and Zhang, Hongbo and Liu, Bingbing and Chen, Ying-Cong},
  journal={arXiv preprint arXiv:2409.18124},
  year={2024}
}

@article{he2024disenvisioner,
  title={DisEnvisioner: Disentangled and Enriched Visual Prompt for Customized Image Generation},
  author={He, Jing and Li, Haodong and Hu, Yongzhe and Shen, Guibao and Cai, Yingjie and Qiu, Weichao and Chen, Ying-Cong},
  journal={arXiv preprint arXiv:2410.02067},
  year={2024}
}

@article{he2025lotus,
  title={Lotus-2: Advancing Geometric Dense Prediction with Powerful Image Generative Model},
  author={He, Jing and Li, Haodong and Sheng, Mingzhi and Chen, Ying-Cong},
  journal={arXiv preprint arXiv:2512.01030},
  year={2025}
}

@article{ho2020denoising,
  title={Denoising diffusion probabilistic models},
  author={Ho, Jonathan and Jain, Ajay and Abbeel, Pieter},
  journal={Advances in neural information processing systems},
  volume={33},
  pages={6840--6851},
  year={2020}
}

@inproceedings{rombach2022high,
  title={High-resolution image synthesis with latent diffusion models},
  author={Rombach, Robin and Blattmann, Andreas and Lorenz, Dominik and Esser, Patrick and Ommer, Bj{\"o}rn},
  booktitle={Proceedings of the IEEE/CVF conference on computer vision and pattern recognition},
  pages={10684--10695},
  year={2022}
}

@article{song2020denoising,
  title={Denoising diffusion implicit models},
  author={Song, Jiaming and Meng, Chenlin and Ermon, Stefano},
  journal={arXiv preprint arXiv:2010.02502},
  year={2020}
}

@article{guo2025long,
  title={Long context tuning for video generation},
  author={Guo, Yuwei and Yang, Ceyuan and Yang, Ziyan and Ma, Zhibei and Lin, Zhijie and Yang, Zhenheng and Lin, Dahua and Jiang, Lu},
  journal={arXiv preprint arXiv:2503.10589},
  year={2025}
}

@article{deng2024autoregressive,
  title={Autoregressive video generation without vector quantization},
  author={Deng, Haoge and Pan, Ting and Diao, Haiwen and Luo, Zhengxiong and Cui, Yufeng and Lu, Huchuan and Shan, Shiguang and Qi, Yonggang and Wang, Xinlong},
  journal={arXiv preprint arXiv:2412.14169},
  year={2024}
}

@article{lamb2016professor,
  title={Professor forcing: A new algorithm for training recurrent networks},
  author={Lamb, Alex M and ALIAS PARTH GOYAL, Anirudh Goyal and Zhang, Ying and Zhang, Saizheng and Courville, Aaron C and Bengio, Yoshua},
  journal={Advances in neural information processing systems},
  volume={29},
  year={2016}
}

@article{bengio2015scheduled,
  title={Scheduled sampling for sequence prediction with recurrent neural networks},
  author={Bengio, Samy and Vinyals, Oriol and Jaitly, Navdeep and Shazeer, Noam},
  journal={Advances in neural information processing systems},
  volume={28},
  year={2015}
}

@misc{mcqueen2008hunger,
  title={Hunger},
  author={Steve McQueen},
  year={2008},
  howpublished = {\url{https://en.wikipedia.org/wiki/Hunger_(2008_film)}}
}

@misc{kubrick1980shining,
  title={The Shining},
  author={Stanley Kubrick},
  year={1980},
  howpublished = {\url{https://en.wikipedia.org/wiki/The_Shining_(film)}}
}

@article{yesiltepe2025infinity,
  title={Infinity-RoPE: Action-Controllable Infinite Video Generation Emerges From Autoregressive Self-Rollout},
  author={Yesiltepe, Hidir and Meral, Tuna Han Salih and Akan, Adil Kaan and Oktay, Kaan and Yanardag, Pinar},
  journal={arXiv preprint arXiv:2511.20649},
  year={2025}
}

@article{zhou2026videomemory,
  title={VideoMemory: Toward Consistent Video Generation via Memory Integration},
  author={Zhou, Jinsong and Du, Yihua and Xu, Xinli and Wang, Luozhou and Zhuang, Zijie and Zhang, Yehang and Li, Shuaibo and Hu, Xiaojun and Su, Bolan and Chen, Ying-cong},
  journal={arXiv preprint arXiv:2601.03655},
  year={2026}
}

@article{zhu2026causal,
  title={Causal forcing: Autoregressive diffusion distillation done right for high-quality real-time interactive video generation},
  author={Zhu, Hongzhou and Zhao, Min and He, Guande and Su, Hang and Li, Chongxuan and Zhu, Jun},
  journal={arXiv preprint arXiv:2602.02214},
  year={2026}
}

@article{kim2026memrope,
  title={Memrope: Training-free infinite video generation via evolving memory tokens},
  author={Kim, Youngrae and Hu, Qixin and Kuo, C-C Jay and Beerel, Peter A},
  journal={arXiv preprint arXiv:2603.12513},
  year={2026}
}

@article{mao2026packforcing,
  title={Packforcing: Short video training suffices for long video sampling and long context inference},
  author={Mao, Xiaofeng and Rui, Shaohao and Ying, Kaining and Zheng, Bo and Li, Chuanhao and Chi, Mingmin and Zhang, Kaipeng},
  journal={arXiv preprint arXiv:2603.25730},
  year={2026}
}

@article{li2026train,
  title={Train short, inference long: Training-free horizon extension for autoregressive video generation},
  author={Li, Jia and Fu, Xiaomeng and Peng, Xurui and Chen, Weifeng and Zheng, Youwei and Zhao, Tianyu and Wang, Jiexi and Chen, Fangmin and Wang, Xing and So, Hayden Kwok-Hay},
  journal={arXiv preprint arXiv:2602.14027},
  year={2026}
}

@article{tian2026head,
  title={Head Forcing: Long Autoregressive Video Generation via Head Heterogeneity},
  author={Tian, Jiahao and Wang, Yiwei and Yu, Gang and Zhang, Chi},
  journal={arXiv preprint arXiv:2605.14487},
  year={2026}
}

@article{wu2026echo,
  title={Echo-Forcing: A Scene Memory Framework for Interactive Long Video Generation},
  author={Wu, Mingqiang and Feng, Weilun and Zhang, Zhefeng and Qin, Haotong and Li, Yuqi and Fan, Guoxin and Liu, Xiaokun and An, Zhulin and Huang, Libo and Xu, Yongjun and others},
  journal={arXiv preprint arXiv:2605.16003},
  year={2026}
}

@article{seedance2026seedance,
  title={Seedance 2.0: Advancing video generation for world complexity},
  author={Seedance, Team and Chen, De and Chen, Liyang and Chen, Xin and Chen, Ying and Chen, Zhuo and Chen, Zhuowei and Cheng, Feng and Cheng, Tianheng and Cheng, Yufeng and others},
  journal={arXiv preprint arXiv:2604.14148},
  year={2026}
}

@inproceedings{zhang2026dvd,
  title={Dvd: Deterministic video depth estimation with generative priors},
  author={Zhang, Hongfei and Chen, Harold Haodong and Liao, Chenfei and He, Jing and Zhang, Zixin and Li, Haodong and Liang, Yihao and Chen, Kanghao and Ren, Bin and Zheng, Xu and others},
  booktitle={ICML 2026 Workshop on Foundations of Deep Generative Models: Understanding Memorization, Generalization, and Reasoning},
  year={2026}
}

@article{oquab2023dinov2,
  title={Dinov2: Learning robust visual features without supervision},
  author={Oquab, Maxime and Darcet, Timoth{\'e}e and Moutakanni, Th{\'e}o and Vo, Huy and Szafraniec, Marc and Khalidov, Vasil and Fernandez, Pierre and Haziza, Daniel and Massa, Francisco and El-Nouby, Alaaeldin and others},
  journal={arXiv preprint arXiv:2304.07193},
  year={2023}
}

@inproceedings{radford2021learning,
  title={Learning transferable visual models from natural language supervision},
  author={Radford, Alec and Kim, Jong Wook and Hallacy, Chris and Ramesh, Aditya and Goh, Gabriel and Agarwal, Sandhini and Sastry, Girish and Askell, Amanda and Mishkin, Pamela and Clark, Jack and others},
  booktitle={International conference on machine learning},
  pages={8748--8763},
  year={2021},
  organization={PmLR}
}
\bibliographystyle{iclr2027_conference}

\clearpage
\appendix
\renewcommand{\thefigure}{S\arabic{figure}}
\renewcommand{\theequation}{S\arabic{equation}}
\renewcommand{\thetable}{S\arabic{table}}
\renewcommand{\thealgorithm}{S\arabic{algorithm}}
\setcounter{figure}{0}
\setcounter{equation}{0}
\setcounter{table}{0}
\setcounter{algorithm}{0}

\noindent\textbf{Supplementary Roadmap.}
App.~\ref{supp:sec:30min} shows the 30-minute rollout above.
App.~\ref{supp:sec:fps} details the efficiency study behind the main-paper FPS numbers.
App.~\ref{supp:sec:abl} reports the full per-dimension ablations under every sink ratio.
App.~\ref{supp:sec:rep} discusses return-to-start repetition and the boundaries of Rolling Semantics.
App.~\ref{supp:sec:periodicity} quantifies periodic replay and content progression at the bank period.
App.~\ref{supp:sec:algo} gives the AR generation pseudo-code.
App.~\ref{supp:sec:vbench} covers the \texttt{VBench-Long} protocol and the dimension legend.

\begin{figure*}[!h]
    \centering
    \includegraphics[width=\linewidth]{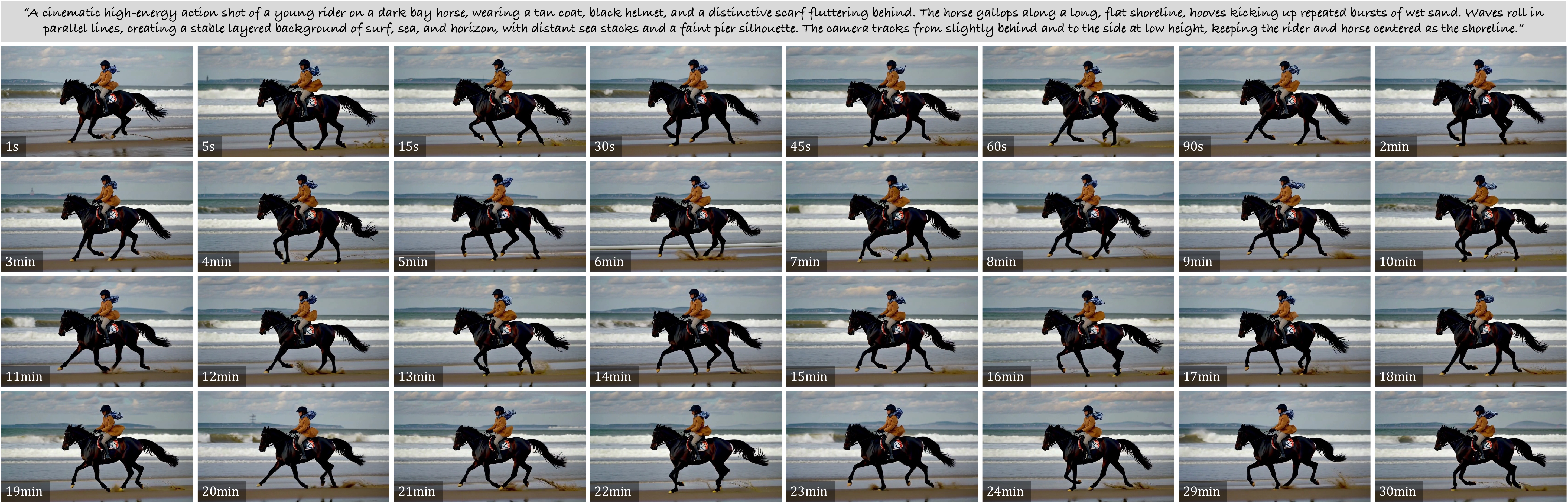}
\vspace{-.5cm}
\caption{
\textbf{30-minute AR rollout with Rolling Sink (qualitative).}
Without extra training, Rolling Sink keeps a 5s-trained AR video diffusion model rolling out for 30 minutes ($360\times$ its training horizon) under a bounded cache, here on realistic and animated content.
Across the shown frames, subject identity, scene structure, colors, and motion stay coherent.
This is a qualitative example: it shows the policy continues to generate at this length, and does not certify quality at arbitrary duration.
}
    \label{supp:fig:30min}
\vspace{-.3cm}
\end{figure*}

\section{Ultra-Long AR Video Synthesis}
\label{supp:sec:30min}

We test Rolling Sink at 30 minutes to see how far the bounded-cache policy runs past the 1- and 5-minute settings evaluated in the main paper.
Fig.~\ref{supp:fig:30min} shows the rollout continuing on both realistic and animated content, with stable subject identity, coherent structure and color, and smooth motion across the shown frames.
We present this length as a qualitative demonstration only.
Automatic long-video metrics and human studies at 30 minutes are open, and fine-grained details can still degrade over such rollouts (Sec.~\ref{sec:con}).

\section{Efficiency Study}
\label{supp:sec:fps}

This section details the FPS numbers reported in Sec.~\ref{sec:exp}.
Rolling Sink keeps the base model's bounded cache budget and few-step denoising, and only changes which cached chunks are selected and how they are re-indexed.
We measure throughput against Self Forcing~\citep{huang2025self} on a single NVIDIA RTX 4090 GPU, under the 1-minute setting of Tab.~\ref{tab:1min}: 961-frame generation, 480$\times$832 resolution, 4 denoising steps, and a 6-chunk cache, excluding I/O.
Tab.~\ref{supp:tab:efficiency} reports 5.35 FPS for Rolling Sink against 5.38 FPS for Self Forcing, a 0.56\% change.
Because the cache holds a fixed $K=6$ chunks, peak GPU memory stays constant as the rollout grows, unlike full-context retention whose memory scales with the number of generated chunks.

\setlength{\cmidrulewidth}{\lightrulewidth}

\begin{table}[!h]
\centering
\scriptsize
\caption{
\textbf{Efficiency comparison.}
Rolling Sink introduces negligible runtime overhead under the same cache budget and denoising setting.
}
\vspace{-.2cm}
\label{supp:tab:efficiency}
\begin{tabular}{lc}
\toprule
Method & FPS $\uparrow$ \\
\midrule
Self Forcing & 5.38 \\
Rolling Sink & 5.35 \\
\bottomrule
\end{tabular}
\vspace{-.2cm}
\end{table}

\section{Quantitative Analysis of Rolling Sink}
\label{supp:sec:abl}

\setlength{\cmidrulewidth}{\lightrulewidth}

\begin{table*}[t]
\centering
\caption{
\textbf{Full ablation on 1-minute AR video synthesis (all \texttt{VBench-Long} dimensions).}
For each sink/cache ratio, we add Attention Sink, Temporal Re-index, and Rolling Semantics in turn, recovering low-drift content, sliding temporal indices, and evolving semantics.
The overall score rises as each behavior is added, matching the main-paper analysis (Fig.~\ref{fig:abl}).
\textbf{Bold} and \underline{underline} mark the best and second-best within each ratio.
}
\vspace{-.2cm}
\resizebox{\textwidth}{!}{%
\label{supp:tab:abl_1min}
\begin{tabular}{lccccccccccccccccc}
\toprule
\multirow{2}{*}{Cache Strategy}
& \multicolumn{16}{c}{\texttt{VBench-Long} Dimensions $\uparrow$}
& Avg \\
\cmidrule(lr){2-17}
& SubCon & BgCon & AesQl & ImgQl & ObjCls & MulObj & Col & SpaRel & Sce & TmpSty & OvrCon & HumAct & TmpFli & MoSmo & DyDg & AppSty & Score$\uparrow$ \\
\midrule
\rowcolor{GroupGray}
\multicolumn{18}{c}{\emph{Sink/Cache Ratio: 1/6}} \\
No Sink Chunk & 96.79 & 96.53 & 59.16 & 69.80 & 86.80 & 36.39 & 64.33 & 71.21 & 10.79 & 22.20 & 19.91 & 68.86 & 97.63 & \underline{98.14} & 48.57 & \textbf{20.99} & 60.51 \\
\textbf{w/} Attention Sink & \textbf{98.70} & 96.61 & \underline{61.21} & \underline{70.04} & 92.91 & \underline{58.84} & 68.36 & 91.78 & 15.87 & 21.91 & 21.36 & \underline{72.00} & \textbf{98.20} & \textbf{99.03} & 16.79 & 20.39 & 62.75 \\
\textbf{w/} Temporal Reindex & 97.62 & \underline{96.67} & 60.80 & \textbf{70.58} & \underline{99.88} & 54.57 & \textbf{75.89} & \underline{97.11} & \underline{17.14} & \underline{22.28} & \underline{22.45} & 68.00 & 97.87 & 97.60 & \underline{67.50} & \underline{20.45} & \underline{66.65} \\
\textbf{w/} Rolling Semantics & \underline{97.66} & \textbf{96.78} & \textbf{61.52} & 68.86 & \textbf{100.00} & \textbf{63.88} & \underline{75.26} & \textbf{98.87} & \textbf{23.49} & \textbf{24.46} & \textbf{23.08} & \textbf{77.14} & \underline{98.18} & 97.64 & \textbf{68.93} & 20.37 & \textbf{68.51} \\
\midrule
\rowcolor{GroupGray}
\multicolumn{18}{c}{\emph{Sink/Cache Ratio: 2/6}} \\
No Sink Chunk & 96.79 & 96.53 & 59.16 & 69.80 & 86.80 & 36.39 & 64.33 & 71.21 & 10.79 & 22.20 & 19.91 & 68.86 & 97.63 & 98.14 & 48.57 & \textbf{20.99} & 60.51 \\
\textbf{w/} Attention Sink & \textbf{98.76} & \textbf{96.91} & \underline{61.87} & \underline{70.12} & 98.50 & \textbf{69.91} & \underline{79.09} & 95.64 & 13.02 & \underline{22.70} & 21.53 & \textbf{83.71} & \underline{98.01} & \textbf{99.12} & 17.14 & 19.95 & 65.37 \\
\textbf{w/} Temporal Reindex & 98.07 & 96.62 & 60.50 & \textbf{70.73} & \underline{99.89} & 63.36 & 78.20 & \underline{96.73} & \textbf{22.86} & 22.47 & \underline{22.01} & 69.14 & 97.80 & 98.06 & \underline{64.64} & \underline{19.98} & \underline{67.57} \\
\textbf{w/} Rolling Semantics & \underline{98.16} & \underline{96.86} & \textbf{62.36} & 69.01 & \textbf{100.00} & \underline{69.37} & \textbf{79.76} & \textbf{99.45} & \underline{21.90} & \textbf{24.77} & \textbf{23.25} & \underline{78.86} & \textbf{98.11} & \underline{98.31} & \textbf{70.36} & 19.69 & \textbf{69.39} \\
\midrule
\rowcolor{GroupGray}
\multicolumn{18}{c}{\emph{Sink/Cache Ratio: 3/6}} \\
No Sink Chunk & 96.79 & 96.53 & 59.16 & \underline{69.80} & \underline{86.80} & 36.39 & 64.33 & 71.21 & 10.79 & 22.20 & 19.91 & 68.86 & 97.63 & 98.14 & 48.57 & \textbf{20.99} & 60.51 \\
\textbf{w/} Attention Sink & \textbf{99.03} & \textbf{96.93} & \underline{62.09} & 69.31 & \textbf{100.00} & \textbf{80.32} & 76.71 & \underline{98.31} & \underline{18.10} & \underline{22.94} & \underline{22.70} & \underline{83.14} & \textbf{98.22} & \textbf{99.14} & 16.43 & 19.36 & 66.42 \\
\textbf{w/} Temporal Reindex & 97.83 & 96.49 & 60.94 & \textbf{69.95} & \textbf{100.00} & 66.93 & \textbf{85.73} & 96.29 & \textbf{21.59} & 22.10 & 22.22 & 72.29 & 97.55 & 97.62 & \textbf{68.57} & 19.50 & \underline{68.47} \\
\textbf{w/} Rolling Semantics & \underline{98.27} & \underline{96.86} & \textbf{62.68} & 69.62 & \textbf{100.00} & \underline{69.93} & \underline{80.52} & \textbf{99.96} & 17.78 & \textbf{24.80} & \textbf{23.04} & \textbf{83.71} & \underline{98.17} & \underline{98.48} & \underline{66.43} & \underline{19.53} & \textbf{69.36} \\
\midrule
\rowcolor{GroupGray}
\multicolumn{18}{c}{\emph{Sink/Cache Ratio: 4/6}} \\
No Sink Chunk & 96.79 & 96.53 & 59.16 & \underline{69.80} & 86.80 & 36.39 & 64.33 & 71.21 & 10.79 & 22.20 & 19.91 & 68.86 & 97.63 & 98.14 & 48.57 & \textbf{20.99} & 60.51 \\
\textbf{w/} Attention Sink & \textbf{99.05} & \textbf{96.92} & \textbf{63.85} & 68.69 & \textbf{100.00} & \textbf{78.21} & \underline{81.93} & \underline{99.88} & \underline{18.41} & \underline{23.72} & \textbf{23.28} & \underline{79.43} & \textbf{98.39} & \textbf{99.16} & 24.29 & 18.90 & 67.13 \\
\textbf{w/} Temporal Reindex & 97.99 & 96.45 & 60.55 & \textbf{70.74} & \underline{99.93} & \underline{70.66} & \textbf{86.75} & 99.65 & 17.46 & 22.96 & \underline{22.28} & 76.86 & 97.62 & 98.22 & \underline{61.79} & \underline{19.21} & \underline{68.69} \\
\textbf{w/} Rolling Semantics & \underline{98.24} & \underline{96.78} & \underline{62.79} & 69.41 & \textbf{100.00} & 69.84 & 80.58 & \textbf{99.99} & \textbf{19.05} & \textbf{24.82} & \textbf{23.28} & \textbf{82.86} & \underline{98.08} & \underline{98.44} & \textbf{70.71} & 19.20 & \textbf{69.63} \\
\midrule
\rowcolor{GroupGray}
\multicolumn{18}{c}{\emph{Sink/Cache Ratio: 5/6}} \\
No Sink Chunk & 96.79 & 96.53 & 59.16 & \underline{69.80} & 86.80 & 36.39 & 64.33 & \underline{71.21} & 10.79 & 22.20 & 19.91 & 68.86 & 97.63 & 98.14 & 48.57 & \textbf{20.99} & 60.51 \\
\textbf{w/} Attention Sink & \textbf{98.98} & \textbf{97.62} & 62.46 & 69.13 & \textbf{100.00} & \textbf{70.00} & \textbf{87.32} & \textbf{100.00} & \underline{23.81} & \textbf{25.11} & \textbf{23.60} & 73.71 & 97.57 & \underline{98.36} & 53.57 & 18.69 & 68.75 \\
\textbf{w/} Temporal Reindex & 98.25 & \underline{97.01} & \underline{62.51} & \textbf{70.39} & \underline{99.96} & 69.96 & 79.36 & \textbf{100.00} & \textbf{26.03} & 24.33 & 23.09 & \textbf{80.00} & \underline{97.66} & 98.26 & \underline{66.07} & 18.87 & \underline{69.48} \\
\textbf{w/} Rolling Semantics & \underline{98.58} & 96.94 & \textbf{63.08} & 69.68 & \textbf{100.00} & \underline{69.98} & \underline{80.23} & \textbf{100.00} & 21.59 & \underline{25.03} & \underline{23.16} & \underline{78.00} & \textbf{98.16} & \textbf{98.65} & \textbf{74.69} & \underline{18.91} & \textbf{69.79} \\
\bottomrule
\end{tabular}%
}
\end{table*}
\setlength{\cmidrulewidth}{\lightrulewidth}

\begin{table*}[t]
\centering
\caption{
\textbf{Full ablation on 5-minute AR video synthesis (all \texttt{VBench-Long} dimensions).}
The longer 5-minute rollout widens the horizon gap, so the effect of each cache behavior is more visible than at 1 minute.
The overall score rises as low-drift content, sliding temporal indices, and evolving semantics are added, with the best result at Rolling Sink, $S/K=5/6$.
\textbf{Bold} and \underline{underline} mark the best and second-best within each ratio.
}
\vspace{-.2cm}
\resizebox{\textwidth}{!}{%
\label{supp:tab:abl_5min}
\begin{tabular}{lccccccccccccccccc}
\toprule
\multirow{2}{*}{Cache Strategy}
& \multicolumn{16}{c}{\texttt{VBench-Long} Dimensions $\uparrow$}
& Avg \\
\cmidrule(lr){2-17}
& SubCon & BgCon & AesQl & ImgQl & ObjCls & MulObj & Col & SpaRel & Sce & TmpSty & OvrCon & HumAct & TmpFli & MoSmo & DyDg & AppSty & Score$\uparrow$ \\
\midrule
\rowcolor{GroupGray}
\multicolumn{18}{c}{\emph{Sink/Cache Ratio: 1/6}} \\
No Sink Chunk & 94.24 & \textbf{96.10} & 42.89 & 57.01 & 33.39 & 14.27 & 61.05 & 35.70 & 4.30 & 11.32 & 11.39 & 27.10 & 98.20 & \underline{98.65} & 24.19 & \underline{20.53} & 45.65 \\
\textbf{w/} Attention Sink & \textbf{95.86} & \underline{95.71} & \underline{50.37} & \underline{61.93} & 52.54 & 23.53 & 52.78 & 45.78 & 5.73 & \underline{14.25} & 16.36 & 44.52 & 96.94 & \textbf{98.88} & 16.94 & 20.48 & 49.54 \\
\textbf{w/} Temporal Reindex & 95.14 & \underline{95.71} & 49.61 & 61.36 & \underline{61.47} & \underline{31.87} & \underline{63.49} & \underline{68.28} & \underline{7.89} & 13.90 & \underline{16.92} & \underline{48.39} & \underline{98.22} & 97.31 & \textbf{70.97} & 20.50 & \underline{56.31} \\
\textbf{w/} Rolling Semantics & \underline{95.35} & 95.23 & \textbf{58.10} & \textbf{64.50} & \textbf{89.35} & \textbf{50.50} & \textbf{69.45} & \textbf{89.40} & \textbf{20.79} & \textbf{23.08} & \textbf{21.85} & \textbf{73.23} & \textbf{98.41} & 98.05 & \underline{54.84} & \textbf{21.20} & \textbf{63.96} \\
\midrule
\rowcolor{GroupGray}
\multicolumn{18}{c}{\emph{Sink/Cache Ratio: 2/6}} \\
No Sink Chunk & 94.24 & \textbf{96.10} & 42.89 & 57.01 & 33.39 & 14.27 & 61.05 & 35.70 & 4.30 & 11.32 & 11.39 & 27.10 & \underline{98.20} & \underline{98.65} & 24.19 & 20.53 & 45.65 \\
\textbf{w/} Attention Sink & \textbf{96.65} & \underline{95.91} & \underline{54.91} & \textbf{69.28} & \underline{81.67} & \underline{43.63} & \underline{72.87} & 68.50 & \underline{8.96} & 15.65 & 17.59 & \underline{65.48} & 98.19 & \textbf{99.05} & 6.05 & \textbf{20.68} & 57.19 \\
\textbf{w/} Temporal Reindex & 95.87 & 95.30 & 51.93 & 62.81 & 81.07 & 39.58 & 68.26 & \underline{80.62} & 7.17 & \underline{16.20} & \underline{17.93} & \underline{65.48} & 97.90 & 97.58 & \underline{64.52} & 20.55 & \underline{60.17} \\
\textbf{w/} Rolling Semantics & \underline{96.45} & 95.66 & \textbf{60.65} & \underline{65.38} & \textbf{98.41} & \textbf{63.79} & \textbf{73.24} & \textbf{96.46} & \textbf{23.30} & \textbf{24.24} & \textbf{22.00} & \textbf{76.77} & \textbf{98.27} & 98.35 & \textbf{68.15} & \underline{20.58} & \textbf{67.61} \\
\midrule
\rowcolor{GroupGray}
\multicolumn{18}{c}{\emph{Sink/Cache Ratio: 3/6}} \\
No Sink Chunk & 94.24 & \textbf{96.10} & 42.89 & 57.01 & 33.39 & 14.27 & 61.05 & 35.70 & 4.30 & 11.32 & 11.39 & 27.10 & 98.20 & \underline{98.65} & 24.19 & \textbf{20.53} & 45.65 \\
\textbf{w/} Attention Sink & \textbf{97.84} & 95.92 & \underline{58.05} & \textbf{69.86} & \underline{95.85} & \textbf{76.11} & 70.07 & \underline{96.03} & \underline{14.70} & \underline{19.59} & \underline{20.11} & \underline{77.10} & \textbf{98.53} & \textbf{99.10} & 14.11 & 20.18 & 63.95 \\
\textbf{w/} Temporal Reindex & 96.10 & 95.13 & 56.57 & \underline{68.79} & 93.71 & 51.75 & \underline{76.67} & 89.70 & 11.47 & 19.43 & 19.99 & 65.48 & 97.62 & 97.03 & \textbf{71.77} & \underline{20.28} & \underline{64.47} \\
\textbf{w/} Rolling Semantics & \underline{96.79} & \underline{95.97} & \textbf{62.06} & 67.05 & \textbf{99.80} & \underline{68.87} & \textbf{80.92} & \textbf{99.87} & \textbf{16.13} & \textbf{24.68} & \textbf{22.37} & \textbf{81.94} & \underline{98.28} & 98.36 & \underline{67.34} & 20.02 & \textbf{68.78} \\
\midrule
\rowcolor{GroupGray}
\multicolumn{18}{c}{\emph{Sink/Cache Ratio: 4/6}} \\
No Sink Chunk & 94.24 & \textbf{96.10} & 42.89 & 57.01 & 33.39 & 14.27 & 61.05 & 35.70 & 4.30 & 11.32 & 11.39 & 27.10 & 98.20 & \underline{98.65} & 24.19 & \textbf{20.53} & 45.65 \\
\textbf{w/} Attention Sink & \textbf{98.47} & 96.02 & \underline{62.10} & \underline{69.51} & \textbf{100.00} & \textbf{79.68} & 76.37 & \underline{99.80} & \textbf{20.43} & \underline{22.58} & \underline{22.19} & 77.74 & \textbf{98.57} & \textbf{99.17} & 27.42 & 19.06 & 66.82 \\
\textbf{w/} Temporal Reindex & 96.46 & 95.13 & 59.01 & \textbf{69.81} & \underline{88.59} & 65.91 & \underline{80.76} & 94.79 & \textbf{20.43} & 22.12 & 21.78 & \underline{78.71} & 97.67 & 97.55 & \underline{67.74} & 19.56 & \underline{67.25} \\
\textbf{w/} Rolling Semantics & \underline{97.01} & \underline{96.07} & \textbf{62.27} & 68.80 & \textbf{100.00} & \underline{69.68} & \textbf{82.51} & \textbf{99.82} & \underline{17.20} & \textbf{24.84} & \textbf{22.46} & \textbf{85.81} & \underline{98.25} & 98.34 & \textbf{72.18} & \underline{19.73} & \textbf{69.69} \\
\midrule
\rowcolor{GroupGray}
\multicolumn{18}{c}{\emph{Sink/Cache Ratio: 5/6}} \\
No Sink Chunk & 94.24 & 96.10 & 42.89 & 57.01 & 33.39 & 14.27 & 61.05 & 35.70 & 4.30 & 11.32 & 11.39 & 27.10 & \underline{98.20} & \textbf{98.65} & 24.19 & \textbf{20.53} & 45.65 \\
\textbf{w/} Attention Sink & \textbf{98.96} & \textbf{97.29} & \underline{62.70} & 69.18 & \textbf{100.00} & 70.00 & \textbf{88.19} & \textbf{100.00} & \underline{22.94} & \underline{25.12} & \textbf{23.56} & 73.23 & 97.58 & 98.36 & \underline{62.90} & 18.64 & 69.29 \\
\textbf{w/} Temporal Reindex & 97.27 & 96.03 & 61.98 & \textbf{70.21} & \underline{99.17} & \underline{70.40} & \underline{82.92} & \underline{99.72} & \textbf{26.16} & 24.45 & 22.90 & \underline{80.97} & 97.65 & 97.91 & 62.50 & \underline{18.95} & \underline{69.32} \\
\textbf{w/} Rolling Semantics & \underline{98.04} & \underline{96.29} & \textbf{62.96} & \underline{69.87} & \textbf{100.00} & \textbf{72.84} & 78.83 & \textbf{100.00} & \textbf{26.16} & \textbf{25.33} & \underline{23.10} & \textbf{87.10} & \textbf{98.32} & \underline{98.59} & \textbf{64.11} & 18.91 & \textbf{70.03} \\
\bottomrule
\end{tabular}%
}
\vspace{-.2cm}
\end{table*}

This section provides the full numerical results for the cache-behavior analysis in Fig.~\ref{fig:abl} of the main paper.
We evaluate three progressive design steps following Sec.~\ref{sec:key}: restoring low-drift content with Attention Sink (Sec.~\ref{sec:attnsink}), sliding temporal indices with Temporal Re-index (Sec.~\ref{sec:re_idx}), and evolving semantics with Rolling Semantics (Sec.~\ref{sec:roll_sem}).
Results are reported under different sink/cache ratios on both 1-minute (Tab.~\ref{supp:tab:abl_1min}) and 5-minute (Tab.~\ref{supp:tab:abl_5min}) AR video synthesis.

Across sink ratios and video lengths, adding each cache behavior raises the overall \texttt{VBench-Long} score~\citep{huang2023vbench,zheng2025vbench2,huang2025vbench++}, which supports jointly maintaining low-drift content, sliding temporal indices, and evolving semantics.
We use $S/K=5/6$ as the default because it scores best while still keeping one recent chunk as a local-motion anchor.
This sweep varies $S/K$ at the fixed budget $K=6$ shared by both bases. Other budgets and chunk sizes would require differently-trained bases (Sec.~\ref{sec:con}).

\section{Discussion on Video Repetition}
\label{supp:sec:rep}

\begin{figure*}[!t]
    \centering
    \includegraphics[width=\linewidth]{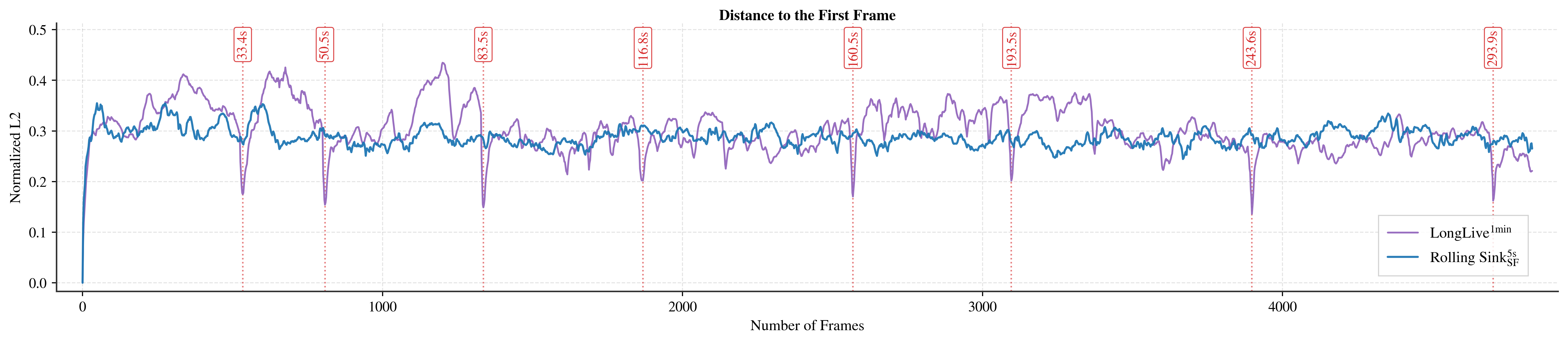}
\vspace{-.5cm}
\caption{
\textbf{Feature distance to the first frame over a long rollout.}
We plot the normalized feature distance between each generated frame and the first frame.
Exact replay of the opening content would drive this distance toward zero in periodic dips.
Against LongLive~\citep{yang2025longlive}, Rolling Sink shows no such collapses at the marked timestamps, which indicates the content keeps changing instead of replaying the start.
This detects one repetition mode (return to the first frame). The bank-periodic replay and progression measure is in App.~\ref{supp:sec:periodicity}.
The marked timestamps are the frames shown in Fig.~\ref{supp:fig:rep}.
}
    \label{supp:fig:rep_curve}
\end{figure*}

\begin{figure*}[!t]
    \centering
    \includegraphics[width=\linewidth]{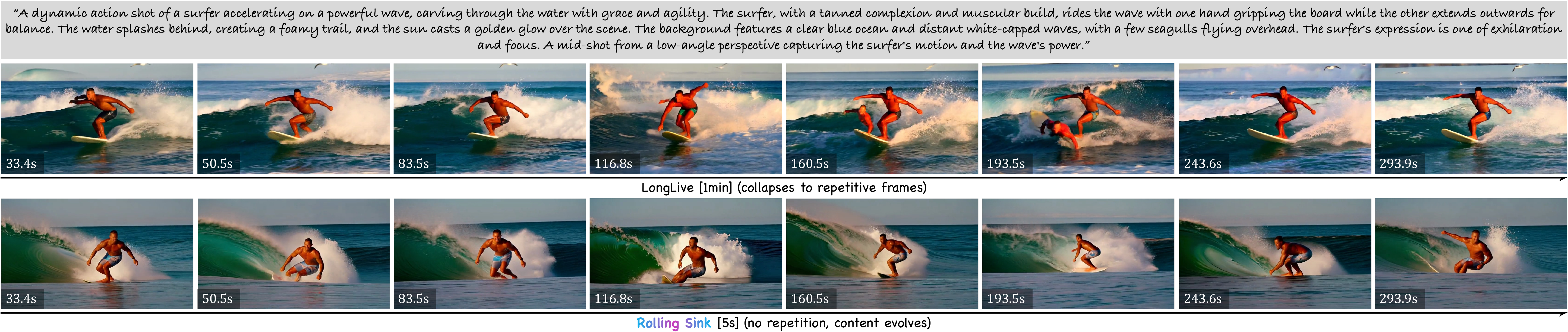}
\vspace{-.5cm}
\caption{
\textbf{Visual examples of content evolution.}
We show frames at the timestamps marked in Fig.~\ref{supp:fig:rep_curve}, which are candidate returns toward earlier visual states.
Against LongLive~\citep{yang2025longlive}, Rolling Sink holds identity, color, and scene structure while poses, motion, camera view, and local appearance keep changing across time, instead of directly replaying earlier frames.
}
    \label{supp:fig:rep}
\vspace{-.3cm}
\end{figure*}

A large sink ratio raises a natural concern: does the model just repeat early content?
Rolling Sink works against this in two ways.
First, it keeps one recent chunk, which carries the current pose, motion phase, camera view, and local appearance, so the next prediction is anchored to the current rollout rather than to early sink content alone.
Second, Rolling Semantics stops the sink from acting as a fixed prefix: it evolves the sink's semantics along a forward-reverse traversal of the low-drift bank and re-indexes the selected chunks to the current temporal window, so both the sink's semantics and positions move with the rollout.

Fig.~\ref{supp:fig:rep_curve} gives diagnostic evidence, comparing Rolling Sink with LongLive~\citep{yang2025longlive}, a training-based baseline trained on 1-minute videos (Tab.~\ref{tab:training_baseline}).
The distance to the first frame stays non-trivial over long rollouts, and the frames in Fig.~\ref{supp:fig:rep} keep evolving while subject and scene structure stay stable.
This diagnostic covers one repetition mode and is not a guarantee against periodicity.
The bank of low-drift chunks is finite, so a long enough rollout can still read out periodically, and the forward-reverse traversal places a partial time-reversal on the sink.
A direct bank-periodic replay and progression measure is given in App.~\ref{supp:sec:periodicity}.
How the forward-reverse schedule affects irreversible motions (an object falling, liquid pouring, something breaking or being consumed) is untested here: we run no with/without Rolling Semantics comparison on such motions and no human study, and mark both open (Sec.~\ref{sec:con}).
The large sink ratio should therefore be read as preserving low-drift cache quality, not as a proof of unbounded content progression.

\section{Periodic Replay and Content Progression}
\label{supp:sec:periodicity}

The finite low-drift bank raises a concern that Rolling Sink could periodically replay or reverse earlier content over long rollouts, rather than letting content progress (Sec.~\ref{sec:con}).
App.~\ref{supp:sec:rep} tracks the distance to the first frame.
Here we measure self-similarity at multiples of the bank length and calibrate it against a bank-loop control.

\noindent\textbf{Setup.}
On five 5-minute rollouts we extract a per-frame feature (DINOv2 ViT-S/14~\citep{oquab2023dinov2} and CLIP ViT-B/32~\citep{radford2021learning} image embedding, $\ell_2$-normalized) and compute the mean cosine distance at time lag $\tau$, $d(\tau)=\mathrm{mean}_t\,[\,1-\langle \boldsymbol{f}_t,\boldsymbol{f}_{t-\tau}\rangle\,]$.
The cache bank spans $K=6$ chunks ($48$ frames) and the forward-reverse schedule completes a cycle every two banks, so replaying or oscillating over the bank would make $d$ dip at bank-multiple lags ($\tau\in\{48,96,144,192\}$) relative to off-bank lags ($\tau\in\{24,72,120,168\}$).
A flat profile (no dip) therefore indicates no measurable bank-periodic replay or reversal.
As a control we tile the first bank ($48$ frames) to full length, a bank-periodic reference.

\setlength{\cmidrulewidth}{\lightrulewidth}
\begin{table}[!t]
\centering
\scriptsize
\caption{
\textbf{Self-similarity at the bank period vs.~off-bank lags} (mean cosine distance over five 5-minute rollouts, larger is more distinct, and $\text{dip}=d_{\text{off-bank}}-d_{\text{bank}}$, where a positive dip demonstrates bank-periodic replay).
A bank loop collapses to $0$ at every bank-multiple lag (large dip), whereas Rolling Sink is uniformly distinct across bank and off-bank lags (dip $\approx0$), so it neither replays nor reverses the finite bank.
}
\vspace{-.2cm}
\label{supp:tab:periodicity}
\begin{tabular}{llccc}
\toprule
Feature & Condition & bank lags & off-bank lags & dip \\
\midrule
\multirow{2}{*}{DINOv2 ViT-S/14}
 & Rolling Sink & $0.043$ & $0.043$ & $-0.00$ \\
 & Bank Loop & $0.000$ & $0.050$ & $\mathbf{+0.05}$ \\
\midrule
\multirow{2}{*}{CLIP ViT-B/32}
 & Rolling Sink & $0.021$ & $0.020$ & $-0.00$ \\
 & Bank Loop & $0.000$ & $0.023$ & $\mathbf{+0.02}$ \\
\bottomrule
\end{tabular}
\end{table}

\noindent\textbf{Result.}
As Tab.~\ref{supp:tab:periodicity} shows, the bank-loop control collapses to zero distance at every bank-multiple lag while staying high off-bank (dip $+0.05$ for DINOv2, $+0.02$ for CLIP), the signature of replaying the bank.
Rolling Sink shows no such dip: its distance at bank-multiple lags matches that at off-bank lags (dip $\approx0$) under both features, so it neither replays the bank nor settles into the two-bank period of the forward-reverse schedule.
Together with the sustained distance to the opening frames in App.~\ref{supp:sec:rep}, this indicates the rollout keeps evolving rather than recycling the finite bank.
This is a global-content periodicity measure.
It does not by itself certify behavior on individual irreversible events (an object falling, liquid pouring, something breaking or being consumed), and long-range novelty is ultimately bounded by the base model's training.
We mark both open in Sec.~\ref{sec:con}.

\section{Algorithm Pseudo-code}
\label{supp:sec:algo}

Alg.~\ref{alg:rolling_sink} summarizes open-ended AR generation with Rolling Sink.
The model first generates the initial $K$ chunks within the training horizon, which serve as a low-drift chunk bank.
After the rollout exceeds the training horizon, Rolling Sink builds the cache at each step by selecting sink chunks from a forward-reverse traversal of the low-drift chunk bank, re-indexing them to the current temporal window, and concatenating them with the recent chunks.
Here, \textsc{Reverse} flips the latent-frame order within a chunk (Eq.~\ref{eq:rs} and~\ref{eq:roll}), and \textsc{ReIndex} applies the RoPE temporal index to each latent frame (Eq.~\ref{eq:tr}).

\begin{algorithm}[t]
\small
\caption{Open-Ended AR Generation with Rolling Sink}
\label{alg:rolling_sink}
\begin{algorithmic}[1]
\Require Base AR video diffusion model $p_\theta$; cache budget $K$; sink size $S$; chunk size $n$; total rollout steps $N$.
\Ensure Generated chunks $\boldsymbol{x}_{[0,N)}$.

\Statex
\State // Generate low-drift chunks within the training horizon
\For{$i = 0$ \textbf{to} $K-1$}
    \State $\mathcal{C}_i \gets \boldsymbol{x}_{[0,i)}$
    \State Sample $\boldsymbol{x}_i \sim p_\theta(\boldsymbol{x}_i \mid \mathcal{C}_i)$
\EndFor

\Statex
\State // Open-ended rollout with Rolling Sink
\For{$i = K$ \textbf{to} $N-1$}
    \State $\mathcal{C}_{\mathrm{sink}} \gets \emptyset$
    \For{$j = i-K$ \textbf{to} $i-(K-S)-1$}
        \State $q \gets \left\lfloor j / K \right\rfloor$, \quad $r \gets j \bmod K$
        \If{$q \bmod 2 = 0$}
            \State $\boldsymbol{s} \gets \boldsymbol{x}_{r}$ \Comment{Forward rolling semantics}
        \Else
            \State $\boldsymbol{s} \gets \textsc{Reverse}\!\left(\boldsymbol{x}_{K-1-r}\right)$ \Comment{Reverse rolling semantics}
        \EndIf
        \State $\boldsymbol{s}^{j} \gets \textsc{ReIndex}\!\left(\boldsymbol{s}, j, n\right)$
        \State $\mathcal{C}_{\mathrm{sink}} \gets \textsc{Append}\!\left(\mathcal{C}_{\mathrm{sink}}, \boldsymbol{s}^{j}\right)$
    \EndFor
    \State $\mathcal{C}_{\mathrm{recent}} \gets \boldsymbol{x}_{[i-(K-S), i)}$
    \State $\mathcal{C}_i \gets \textsc{Cat}\!\left(\mathcal{C}_{\mathrm{sink}}, \mathcal{C}_{\mathrm{recent}}\right)$
    \State Sample $\boldsymbol{x}_i \sim p_\theta(\boldsymbol{x}_i \mid \mathcal{C}_i)$
\EndFor
\State \Return $\boldsymbol{x}_{[0,N)}$
\end{algorithmic}
\end{algorithm}

\section{\texttt{VBench-Long} Evaluation Protocol}
\label{supp:sec:vbench}

\texttt{VBench-Long}~\citep{huang2023vbench,huang2025vbench++,zheng2025vbench2} scores a long video on 16 dimensions, each read out by a separate pretrained expert and all higher-is-better.
Our quantitative tables (Tab.~\ref{tab:1min}, \ref{tab:5min}, \ref{tab:training_baseline}, \ref{supp:tab:abl_1min}, and \ref{supp:tab:abl_5min}) report every dimension under the abbreviations of Tab.~\ref{supp:tab:vbench_legend}.

\noindent\textbf{Prompt suite.}
Each dimension ships with 70 to 100 prompts.
We draw a fixed subset of 10 prompts per dimension once and reuse it for every quantitative result in the paper.

\begin{table}[!t]
\scriptsize
\centering
\caption{\textbf{Abbreviation legend for the \texttt{VBench-Long} dimensions} reported in the quantitative tables, listed in column order.}
\label{supp:tab:vbench_legend}
\vspace{-.2cm}
\begin{tabular}{llll}
\toprule
Abbr. & Dimension & Abbr. & Dimension \\
\midrule
SubCon & subject consistency    & Sce    & scene \\
BgCon  & background consistency  & TmpSty & temporal style \\
AesQl  & aesthetic quality       & OvrCon & overall consistency \\
ImgQl  & imaging quality         & HumAct & human action \\
ObjCls & object class            & TmpFli & temporal flickering \\
MulObj & multiple objects        & MoSmo  & motion smoothness \\
Col    & color                   & DyDg   & dynamic degree \\
SpaRel & spatial relationship    & AppSty & appearance style \\
\bottomrule
\end{tabular}
\vspace{-.2cm}
\end{table}

\end{document}